\documentclass[preprint,12pt,authoryear]{elsarticle}

\usepackage{hyperref}
\hypersetup{
    colorlinks=true,
    linkcolor=blue,
    citecolor=blue,
    urlcolor=blue
}
\usepackage{amssymb}
\usepackage{amsmath}
\usepackage{natbib}
\usepackage{tcolorbox}
\usepackage{tikz}
\usepackage{pgfplots}
\usepackage{tabularray}
\usepackage{float}
\usepackage{tabularx}
\usepackage{multirow}
\usepackage{booktabs}
\usepackage{import}
\usepackage{comment}
\usepackage{orcidlink}
\usepackage{booktabs}
\usepackage{pifont}
\usepackage{rotating}
\usepackage{array}
\usepackage{geometry}
\usepackage{fontawesome5}
\usetikzlibrary{calc, positioning}
\usepackage{lineno}
\usepackage{geometry}
\usepackage{tikz}
\usetikzlibrary{positioning, calc, shadows.blur, shapes.geometric}
\usepackage{lmodern}
\usepackage{tabularx}
\usepackage{booktabs}
\usepackage{multirow}
\usepackage{enumitem}
\usepackage{graphicx}
\usepackage{makecell}
\usepackage{adjustbox}

\definecolor{ic1col}{HTML}{2A527A} 
\definecolor{ic2col}{HTML}{319A9A} 
\definecolor{ic3col}{HTML}{71A543} 
\definecolor{ic4col}{HTML}{E5B626} 
\definecolor{ic5col}{HTML}{E27A2E} 
\definecolor{ic6col}{HTML}{E05658} 
\definecolor{ic7col}{HTML}{7F58A6} 
\definecolor{ic8col}{HTML}{5D5698} 

\newcommand{\cmark}{\ding{51}}
\newcommand{\xmark}{\ding{55}}

\usetikzlibrary{arrows.meta, positioning, shapes.misc, backgrounds}
\usepackage{enumitem}

\usepackage[table]{xcolor}
\usepackage{siunitx}

\journal{Intelligent Systems with Applications}

\begin{document}

\begin{frontmatter}

\title{Mathematical Reasoning in Large Language Models: Benchmarks, Architectures, Evaluation, and Open Challenges}
\author[NUST]{Husnain Amjad}
\ead{hamjad.msds23seecs@seecs.edu.pk}

\author[MIUN]{Raja Khurram Shahzad\orcidlink{0000-0003-2806-9694}}
\ead{raja-khurram.shahzad@miun.se}

\author[West]{Aamir Shahzad\orcidlink{0000-0003-1730-3609}}
\ead{shahzad.aamir@gmail.com}

\author[NUST]{Mehwish Fatima\orcidlink{0000-0003-3424-2991}}
\ead{mehwish.fatima@seecs.edu.pk}

%

\affiliation[NUST]{organization={School of Electrical Engineering and Computer Science, National University of Sciences and Technology, Islamabad, Pakistan}}

\affiliation[West]{organization={School of Computing, Data, and Mathematical Sciences, Western Sydney University, Indonesia}}

\affiliation[MIUN]{organization={Department of Communication, Quality Management and Information Systems, Mid Sweden University, Östersund Campus, Sweden}}

\begin{abstract}
Mathematical reasoning is one of the most challenging tasks for Large Language Models (LLMs) due to their inherent nature of multi-step inference, symbolic interpretation, and accurate logical consistency. Recently, this field has emerged as a new research problem resulting in diverse datasets, training paradigms, reasoning frameworks, and evaluation protocols, making it difficult to determine which methodological advances genuinely improve mathematical reasoning and which primarily exploit benchmark characteristics. We present a structured analysis of approximately 120 peer-reviewed and pre-print studies of mathematical reasoning from a systems perspective. We organize the literature across the reasoning pipeline, dataset design, architectural advances, training and reasoning-enhancement strategies, and evaluation methodologies. We introduce a functional taxonomy of mathematical datasets that distinguishes pretraining corpora, supervised fine-tuning resources, and evaluation benchmarks according to their roles in model development and reasoning complexity. We further analyze how architectural advancements, tool integration, verifier-guided reasoning, reinforcement learning, and parameter-efficient fine-tuning influence reasoning, symbolic competence, and generalization. Our analysis shows that reported improvements in benchmarks frequently arise from better reasoning elicitation, symbolic execution, or verification rather than stronger intrinsic reasoning capabilities. Our analysis further identifies recurring limitations, including benchmark contamination, synthetic reasoning bias, unfaithful reasoning traces, weak symbolic grounding, and the disconnect between answer-level accuracy and process-level correctness. Based on these findings, we argue that mathematical reasoning should be viewed as a systems-level challenge spanning data, models, training, inference, and evaluation rather than as a property of individual models or prompting strategies. We conclude our work by outlining the methodological advances required for developing reliable mathematical reasoning systems, including faithful reasoning mechanisms, robust process-level evaluation, structure-aware representations, and hybrid neuro-symbolic reasoning.
\end{abstract}

\end{frontmatter}


\section{Introduction}\label{sec1}
Mathematical reasoning is one of the most demanding capabilities of Large Language Models (LLMs) because of its structured symbolic manipulation, multi-step deduction, variable binding, and logically consistent inference instead of merely fluent language generation~\citep{devlin1998language,kline1990mathematical,huang2023towards}. Traditional natural language generation tasks require fluent and coherent text generation that usually does not require global logical consistency. However, mathematical problem solving depends on maintaining global logical consistency across intermediate reasoning steps, where a single incorrect step can invalidate the final solution \citep{yuan2023well}. These requirements make mathematics a rigorous benchmark for evaluating LLMs against systematic reasoning.

Recent LLMs often demonstrate good performance across diverse NLP (Natural Language Processing) tasks through extensive pretraining \citep{wei2022chain,matzakos2023learning}, Chain-of-Thought (CoT) prompting \citep{wei2022chain}, and Tool-Integrated Reasoning (TIR) \citep{li2024enhancing}. These techniques usually improve multi-step problem solving and enable progress in algebraic manipulation, theorem proving, and structured word problems \citep{imani2023mathprompter}. However, solving mathematical problems remains a challenge in LLMs \citep{feigenbaum1963computers,hosseini2014learning}. Despite these advances, LLMs often struggle with breakable reasoning, lack of generalization, arithmetic inconsistencies (see Figure~\ref{fig:fig_1}), symbolic reasoning failures, and error propagation across long reasoning chains \citep{rae2021scaling,lewkowycz2022solving}. 

\begin{figure}[t]
    \centering
    \caption{\textbf{Top:} Math word problem. \textbf{Bottom:} Step-by-step erroneous solution.}
    \label{fig:fig_1}
    \scalebox{1}{
    \begin{tcolorbox}[colback=white, colframe=blue!30, colbacktitle=blue!20, coltitle=black, title=Input,width=\linewidth, fontupper=\scriptsize, fonttitle=\scriptsize]
    
    Question:\\
    Dane's two daughters need to plant a certain number of flowers each to grow a garden. As the days passed, the flowers grew into 20 more, but 10 of them died. Dane's daughters harvested the flowers and split them between 5 different baskets, with 4 flowers in each basket. How many flowers did each daughter plant initially?\\
    
    Answer:\\
    
    \end{tcolorbox}
    }
    \scalebox{1}{
    \begin{tcolorbox}[colback=white, colframe=blue!30, colbacktitle=blue!20, coltitle=black, title=Step-by-Step Solution,width=\linewidth, fontupper=\scriptsize, fonttitle=\scriptsize,  before upper={
    \setlength{\abovedisplayskip}{4pt}
    \setlength{\belowdisplayskip}{4pt}
    \setlength{\abovedisplayshortskip}{2pt}
    \setlength{\belowdisplayshortskip}{2pt}
  }]
    
    Calculate the total number of flowers harvested: Dane's daughters split the flowers into 5 baskets, with 4 flowers in each basket. The total number of flowers harvested is: 
    \[5 \times 4 = 20 \ \text{flowers}\]
    Account for the flowers that grew and died: The problem states that 20 additional flowers grew, but 10 of them died. Let the initial number of flowers planted by the two daughters be \(x\). After accounting for the growth and death of flowers, the total number of flowers becomes: 
    \[x + 20 - 10 = x + 10\]
    Set up the equation for the total flowers harvested: The total number of flowers harvested is equal to the total number of flowers after growth and death:  
    \[x + 10 = 20\]
    Solve for \(x\): Subtract 10 from both sides:  
    \[x = 10\]
    Determine how many flowers each daughter planted: The two daughters together planted \(x = 10\) flowers. Since the flowers were planted equally between the two daughters, each daughter planted:  
    \[\frac{10}{2} = 5 \ \text{flowers}\]
    Final Answer:  
    6
    \end{tcolorbox}
    }
\end{figure}

Figure~\ref{fig:fig_1} illustrates a representative example where the model correctly decomposes the problem, derives the intermediate equation $x+10=20$, and solves $x=10$, yet still produces an incorrect final answer. The failure does not arise from an inability to generate plausible reasoning steps but from the absence of global consistency between intermediate derivations and the final prediction. Therefore, LLM performance is substantially compromised on complex mathematical problems such as competition-level mathematics, theorem proving, and university-level problems. All these complex problems require abstraction, symbolic precision, and long-horizon reasoning. This example illustrates that mathematical reasoning failures cannot be attributed to a single modeling deficiency. Instead, they emerge from the interaction of multiple components involved in the reasoning pipeline.

\begin{table}[t]
\centering
\scriptsize
\caption{Comparison of our work with recent surveys on mathematical reasoning in LLMs. The compared surveys are Ahn~\citep{ahn2024large}, Huang~\citep{huang2023towards}, Liu~\citep{liu2023mathematical}, Forootani~\citep{forootani2025survey}, and Zhang~\citep{zhang2026systems}. A \cmark\ indicates explicit coverage, whereas a \xmark\ indicates limited or no coverage.}
\label{tab:survey_comparison}

\renewcommand{\arraystretch}{1.15}
\setlength{\tabcolsep}{4pt}

\resizebox{0.95\linewidth}{!}{%
\begin{tabular}{lcccccc}
\toprule
\textbf{Feature} &
\textbf{Ahn} &
\textbf{Huang} &
\textbf{Liu} &
\textbf{Forootani} &
\textbf{Zhang} &
\textbf{Our Work} \\
\midrule

\multicolumn{7}{l}{\textbf{Datasets}}\\
\midrule
Math focus               & \cmark & \xmark & \cmark & \cmark & \xmark & \cmark \\
Dataset taxonomy         & \xmark & \xmark & \cmark & \xmark & \xmark & \cmark \\
Olympiad benchmarks      & \xmark & \xmark & \cmark & \xmark & \xmark & \cmark \\
Symbolic encoding        & \xmark & \xmark & \cmark & \xmark & \xmark & \cmark \\

\midrule
\multicolumn{7}{l}{\textbf{Model Development}}\\
\midrule
Architectures            & \xmark & \cmark & \cmark & \cmark & \cmark & \cmark \\
Training                 & \xmark & \xmark & \cmark & \cmark & \cmark & \cmark \\
reasoning-enhancement    & \cmark & \cmark & \cmark & \cmark & \cmark & \cmark \\

\midrule
\multicolumn{7}{l}{\textbf{Evaluation}}\\
\midrule
Process evaluation       & \xmark & \xmark & \xmark & \xmark & \xmark & \cmark \\
Faithfulness             & \xmark & \cmark & \xmark & \xmark & \cmark & \cmark \\
Failure modes            & \xmark & \xmark & \xmark & \xmark & \xmark & \cmark \\

\midrule
\multicolumn{7}{l}{\textbf{Analysis}}\\
\midrule
Contamination            & \xmark & \xmark & \xmark & \xmark & \xmark & \cmark \\
Contamination modeling   & \xmark & \xmark & \xmark & \xmark & \xmark & \cmark \\
Memorization             & \xmark & \xmark & \xmark & \xmark & \xmark & \cmark \\
Cross-benchmark analysis & \xmark & \xmark & \xmark & \xmark & \xmark & \cmark \\
Research roadmap         & \xmark & \xmark & \xmark & \xmark & \xmark & \cmark \\

\bottomrule
\end{tabular}
}
\end{table}

These performance failures arise from multiple factors of the reasoning pipeline: (1) Benchmark datasets frequently emphasize final-answer accuracy while remaining susceptible to distributional bias and pretraining contamination, limiting their ability to measure authentic reasoning generalization~\citep{mishra2022lila,huang2025math,guan2025rstar}. (2) In models, auto-regressive next-token prediction optimizes local token probabilities instead of globally valid derivations, encouraging shortcut learning and surface-level reasoning~\citep{ahn2024large}. Although self-reflection, meta-cognitive prompting, and verifier-guided reasoning improve solution quality~\citep{Wang2023MetacognitivePI}, they do not necessarily yield faithful reasoning processes, resulting in generated explanations that remain disconnected from the model's underlying computation~\citep{didolkar2024metacognitive}.

Despite rapid progress in this field, understanding mathematical reasoning remains understudied from a systems perspective. It requires analyzing the interactions among data, model design, training, inference, and evaluation rather than examining these components independently. Improvements in new architectures often depend on the datasets used for supervision; gains from reasoning strategies frequently rely on external tools or verification mechanisms; and benchmark performance is strongly influenced by evaluation protocols. Therefore, these advances cannot be interpreted solely as improvements in intrinsic reasoning capability. Existing surveys examine individual aspects of mathematical reasoning, such as prompting strategies, benchmark datasets, reasoning architectures, or evaluation methodologies. While these surveys provide valuable insights, they do not explain how decisions made across different stages of the model development pipeline jointly impact mathematical reasoning. This independent treatment makes it difficult to distinguish improvements arising from stronger reasoning mechanisms to those resulting from better data, supervision, inference strategies, or evaluation methods. Table~\ref{tab:survey_comparison} presents a comparison of our work with a few recent surveys, emphasizing the broader system perspective adopted in this paper.

\begin{figure}[t]
\centering
\includegraphics[width=\textwidth]{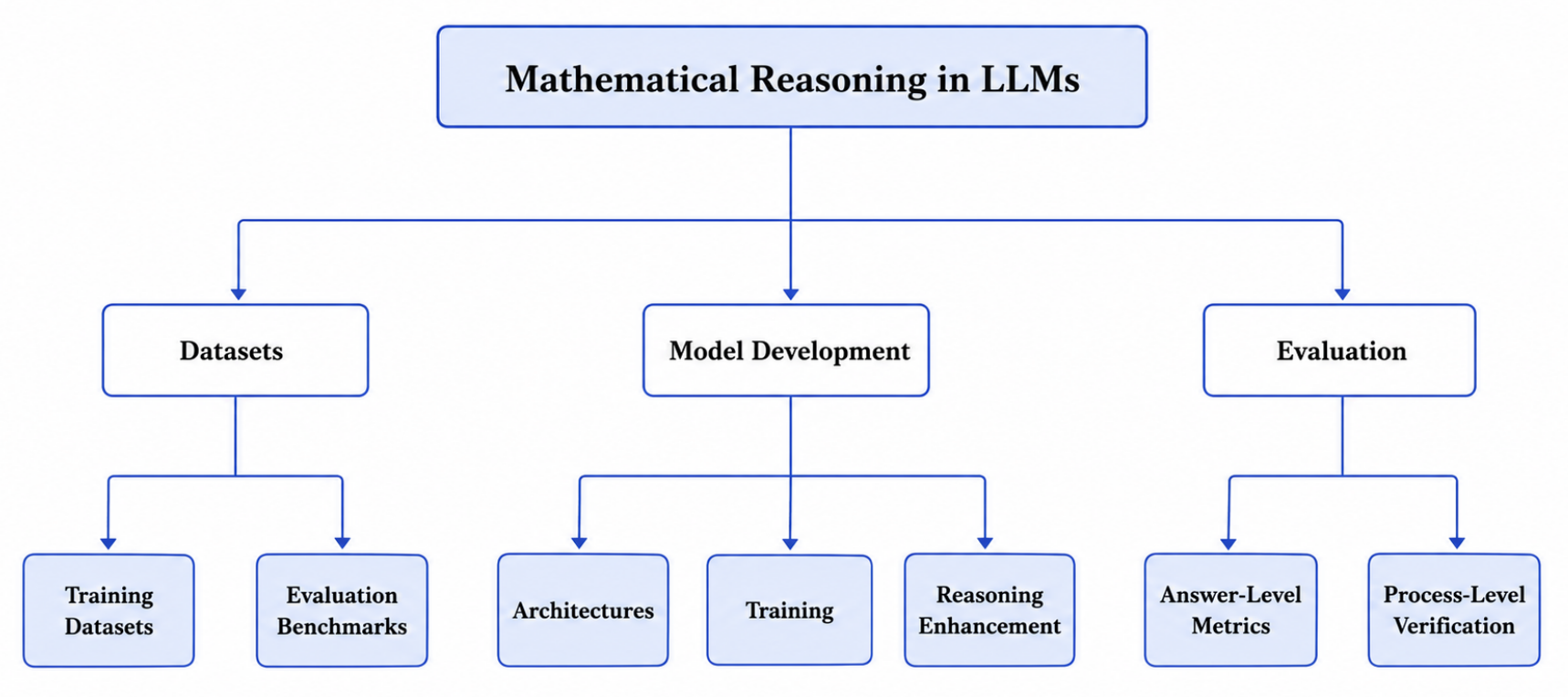}
\caption{System-level organization of mathematical reasoning in LLMs adopted in this survey. We organize the literature around three interconnected components of the reasoning pipeline: datasets, model development, and evaluation. Their interactions collectively determine the reasoning capability, robustness, and reliability of LLM-based mathematical reasoning.}
\label{fig:math_llm_landscape}
\end{figure}

We present a structured review of mathematical reasoning in LLMs from a system-level perspective. Instead of treating datasets, architectures, training strategies, and evaluation as separate topics, we analyze their interactions throughout the entire model development life cycle. We begin by discussing mathematical datasets based on their functional roles in pretraining, supervised fine-tuning, and evaluation, covering foundational corpora, competition mathematics, and formal theorem proving (Section~\ref{sec4}). We then examine architectural advancements, training and reasoning-enhancement strategies (including tool integration, verifier-guided reasoning, reinforcement learning, and parameter-efficient adaptation) to analyze how they influence reasoning robustness, symbolic competence, and generalization (Section~\ref{sec5}). Finally, we evaluate existing evaluation methodologies, highlighting the gap between answer-level accuracy and process-level reasoning verification, and synthesize the challenges that continue to limit reliable mathematical reasoning in LLMs (Section~\ref{sec6}). 

Figure~\ref{fig:math_llm_landscape} summarizes the overall research studies covered in our work and illustrates the relationship between datasets, model development, and evaluation methodologies. This system-level perspective motivates the following research questions, which guide the analysis throughout the remainder of the survey.

\subsection{Research Questions}
We investigate mathematical reasoning in LLMs through the following research questions:

\begin{itemize}[noitemsep, topsep=0pt]
\item \textbf{RQ1:} 
How do existing datasets and benchmarks support the development and evaluation of mathematical reasoning in LLMs, and what aspects of reasoning do they actually measure?

\item \textbf{RQ2:} 
How do architectural adaptations, training paradigms, and reasoning-enhancement strategies influence reasoning and generalization in LLMs?

\item \textbf{RQ3:} 
To what extent do existing LLMs generate faithful and verifiable reasoning processes that are consistent with their final predictions?

\item \textbf{RQ4:} 
How effectively do existing evaluation techniques assess mathematical reasoning, and what limitations affect their reliability?

\item \textbf{RQ5:} 
What are the major challenges limiting reliable mathematical reasoning in LLMs, and which research directions are most critical for advancing the field?

\end{itemize}

\subsection{Contributions}
We make the following contributions:

\begin{itemize}[noitemsep, topsep=0pt]
\item We present a system-level analysis of mathematical reasoning in LLMs.  We analyze the interactions among datasets, architectures, training strategies, and evaluation methodologies across the entire model development life cycle to investigate their collective impact on mathematical reasoning.

\item We introduce a functional taxonomy of mathematical datasets by organizing them according to their role in model development, and evaluation benchmarks across different levels of reasoning complexity.

\item We examine architectural advancements, tool integration, verifier-guided reasoning, and parameter-efficient training for understanding their impact on reasoning robustness and generalization.

\item We analyze existing evaluation frameworks and the gap between answer-level accuracy and process-level reasoning verification, together with their implications for reasoning faithfulness and benchmark reliability.

\item We identify recurring limitations, including benchmark contamination, synthetic reasoning bias, weak symbolic grounding, and unfaithful reasoning, and discuss the methodological advances required for developing reliable mathematical reasoning models.

\end{itemize}

\subsection{Paper Organization}
We organize the remainder of this survey according to the major components of the mathematical reasoning pipeline. Section~\ref{sec2} describes the structured literature review methodology. Section~\ref{sec3} introduces the background on mathematical reasoning in LLMs. Section~\ref{sec4} examines mathematical datasets and benchmarks. Section~\ref{sec5} analyzes model architectures, training paradigms, and reasoning-enhancement strategies. Section~\ref{sec6} reviews evaluation methodologies and discusses their limitations. Section~\ref{sec7} synthesizes the major challenges and future research directions, while Section~\ref{sec9} concludes the survey.


\section{Structured Literature Review Methodology}\label{sec2}
This survey follows the Preferred Reporting Items for Systematic Reviews and Meta-Analyses (PRISMA)\footnote{https://www.prisma-statement.org/} 2020 reporting guidelines to identify, screen, and synthesize research on mathematical reasoning in Large Language Models (LLMs). We conduct a structured literature review because the literature spans diverse datasets, model families, prompting strategies, training procedures, evaluation protocols, and rapidly evolving preprints. These differences make a formal quantitative meta-analysis inappropriate. We use the PRISMA reporting guidelines to document the search process, eligibility criteria, screening decisions, and reproducibility safeguards, rather than to claim exhaustive coverage of a rapidly evolving research area.

We define the review protocol before literature screening to ensure a transparent and reproducible study selection process. We organize the protocol around the five research questions introduced in Section~\ref{sec1}, covering datasets and benchmarks, architectural adaptations and training strategies, reasoning faithfulness, evaluation methodologies, and open challenges related to robustness, contamination, memorization, and generalization. The protocol specifies the information sources, keyword engineering strategy, Boolean query formulation, retrieval procedure, eligibility criteria, evidence extraction, and thematic synthesis adopted throughout this review.

\subsection{Review Design and Scope}
We map, classify, and critically synthesize research on LLM-based mathematical reasoning across the complete model development pipeline. Consistent with the systems perspective introduced in Section~\ref{sec1}, we examine mathematical reasoning through the interaction among datasets, model development, reasoning mechanisms, and evaluation rather than treating these components independently. We cover benchmarks for mathematical reasoning, training and adaptation resources, reasoning architectures, prompting and inference-time reasoning mechanisms, verifier-guided and tool-augmented systems, process-level evaluation, contamination analysis, symbolic representation, and generalization failure modes.  We consider survey and technical synthesis papers to understand research trends and establish comparison baselines, while empirical and methodological studies serve as the primary evidence base for our thematic analysis.

We do not organize the literature solely according to benchmark performance or reported accuracy. Instead, we map each study according to its role within the reasoning pipeline, including data construction, representation, model architecture, training, inference-time reasoning enhancement, verification, evaluation, robustness, and failure analysis. Our organization provides the analytical framework used throughout the remainder of the survey.

\begin{table}[t]
\centering
\small
\caption{Inclusion criteria used for primary study selection.}
\label{tab:inclusion-criteria}
\begin{tabularx}{\linewidth}{p{0.10\linewidth}X}
\toprule
ID & Inclusion criterion \\
\midrule
IC1 & Proposes or evaluates an LLM-based mathematical reasoning method, model, or reasoning pipeline. \\
IC2 & Investigates automated mathematical reasoning, theorem proving, symbolic reasoning, proof generation, or autoformalization using AI/LLMs. \\
IC3 & Introduces, evaluates, or analyzes mathematical reasoning datasets, benchmarks, corpora, or benchmark taxonomies. \\
IC4 & Studies reasoning-oriented training or adaptation, including pretraining, supervised fine-tuning, instruction tuning, reinforcement learning, policy optimization, distillation, or parameter-efficient fine-tuning. \\
IC5 & Analyzes prompting or inference-time reasoning mechanisms such as Chain-of-Thought, self-consistency, tool-integrated reasoning, verifier-guided reasoning, retrieval augmentation, or process supervision. \\
IC6 & Examines answer-level, process-level, symbolic, formal, or hybrid evaluation methodologies for mathematical reasoning. \\
IC7 & Discusses reasoning, faithfulness, hallucination, transparency, interpretability, verification, contamination, memorization, robustness, or out-of-distribution generalization. \\
IC8 & Provides a high-quality survey, benchmark comparison, or technical synthesis needed to contextualize the above themes. \\
\bottomrule
\end{tabularx}
\end{table}

\subsection{Eligibility Criteria}
We determine study eligibility according to each publication's conceptual relevance to LLM-based mathematical reasoning. We retain a publication if it satisfies at least one inclusion criterion and does not meet any exclusion criterion. We apply these criteria consistently during title screening, abstract screening, and full-text evidence extraction.

\begin{table}[t]
\centering
\small
\caption{Exclusion criteria used during screening and eligibility assessment.}
\label{tab:exclusion-criteria}
\begin{tabularx}{\linewidth}{p{0.10\linewidth}X}
\toprule
ID & Exclusion criterion \\
\midrule
EC1 & Generic LLM/NLP studies without explicit mathematical reasoning, theorem proving, symbolic reasoning, or math benchmark content. \\
EC2 & Papers focused only on unrelated applications, deployment issues, or optimization/control without a mathematical reasoning contribution. \\
EC3 & Non-English publications, inaccessible records, or papers lacking sufficient methodological or technical detail for extraction. \\
EC4 & Duplicate records or multiple versions of the same work; the most recent and technically complete version was retained, with published versions preferred when equivalent. \\
EC5 & Opinion pieces or informal commentary without substantive empirical, methodological, dataset, architectural, or evaluation content. \\
\bottomrule
\end{tabularx}
\end{table}

\subsection{Information Sources}
We conducted literature searches on 16 and 17 February 2026 to establish a fixed temporal boundary for the review. We use five information sources: Google Scholar, ACM Digital Library, IEEE Xplore, ScienceDirect, and arXiv. Together, these sources cover peer-reviewed journals, conference proceedings, workshop papers, technical reports, and high-impact preprints in artificial intelligence, machine learning, natural language processing, automated reasoning, and theorem proving.

We record indexed hits at the database level, together with the retrieved records used for screening. Because Google Scholar and several publisher databases restrict bulk exports and dynamically rank search results, we do not treat indexed hits as individually screened records. Instead, we use them to document the breadth of the search, while exported records and relevance-window retrievals form the screened corpus.

\subsection{Keyword Engineering}
We design the keyword strategy to reflect the systems-level organization adopted throughout this survey. Rather than relying on a single search expression, we organize the search around the major components of mathematical reasoning, including datasets, model development, reasoning mechanisms, and evaluation. This strategy reduces retrieval bias stemming from inconsistent terminology across different research communities while ensuring coverage of the topics addressed by the five research questions.

\begin{figure}[t]
\centering
\includegraphics[width=\linewidth]{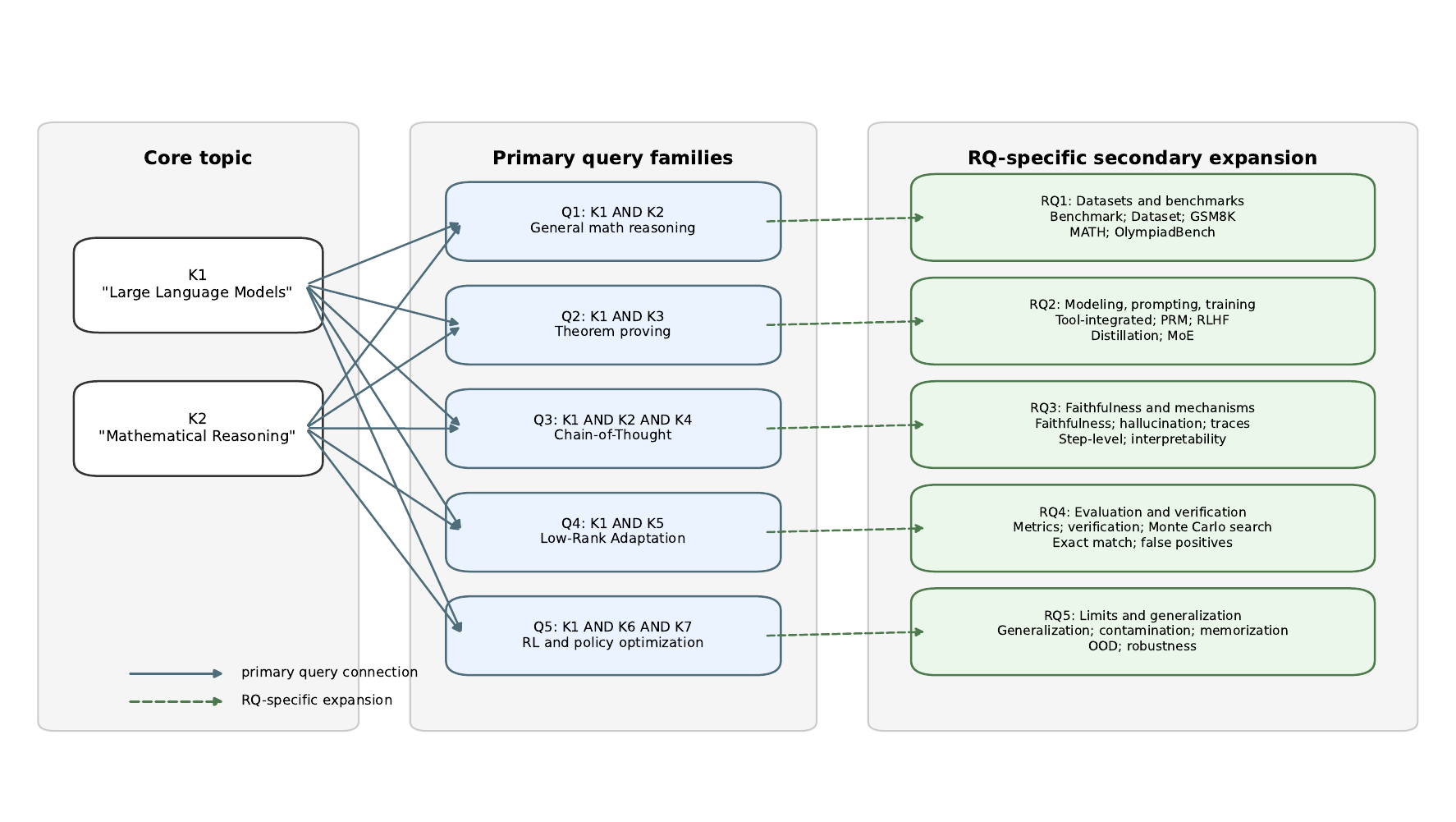}
\caption{Keyword and query architecture aligned with research questions.}
\label{fig:keyword-architecture}
\end{figure}

We adopt a hierarchical keyword strategy with primary and secondary keywords (Figure~\ref{fig:keyword-architecture}). We define seven primary keywords at the first level that outline the conceptual scope of the review: K1 Large Language Models, K2 Mathematical Reasoning, K3 Theorem Proving, K4 Chain-of-Thought, K5 Low-Rank Adaptation, K6 Reinforcement Learning, and K7 Policy Optimization. We organize secondary keywords around five research questions at the second level. These groups capture terminology related to datasets and benchmarks; modeling and training strategies; reasoning faithfulness; evaluation and formal verification; and limitations such as contamination, memorization, robustness, and out-of-distribution generalization. This hierarchical organization enables broad retrieval while preserving targeted recovery of specialized studies that a single search expression may overlook.

\begin{table}[t]
\centering
\small
\caption{Database-level indexed hits record.}
\label{tab:database-counts}
\begin{tabularx}{\linewidth}{p{0.25\linewidth}p{0.14\linewidth}X}
\toprule
Source & Indexed hits & Role in search strategy \\
\midrule
Google Scholar & 95,612 & Logged source; retrieval constrained by ranked-result export limitations. \\
ACM Digital Library & 4,285 & Peer-reviewed computer science and AI venues. \\
IEEE Xplore & 881 & Engineering and AI conference/journal coverage. \\
ScienceDirect & 3,367 & Journal and proceedings coverage in AI, data science, and applied systems. \\
arXiv & 2,755 & High-impact pre-prints in LLMs, reasoning, and theorem proving. \\
Total & 106,900 & Indexed hits before relevance-window extraction. \\
\bottomrule
\end{tabularx}
\end{table}

\subsection{Query Formulation}
We formulate ten Boolean search configurations to retrieve studies within the scope of this survey. Queries Q1 to Q5 define the primary search families derived from the core keywords, whereas Q6 to Q10 extend these queries to address the five research questions. We use the AND operator to require intersections between independent concepts, the OR operator to capture synonyms and semantically equivalent terms, and parentheses to preserve logical precedence across different search engines.

\begin{table}[t]
\centering
\scriptsize
\caption{Search configurations used during literature retrieval.}
\label{tab:boolean-queries}
\begin{tabularx}{\linewidth}{p{0.06\linewidth}p{0.10\linewidth}p{0.22\linewidth}X}
\toprule
ID & Category & Research objective & Boolean query \\
\midrule
Q1 & Primary & General mathematical reasoning & ``Large Language Models'' AND ``Mathematical Reasoning'' \\
Q2 & Primary & Formal mathematical reasoning and theorem proving & ``Large Language Models'' AND ``Theorem Proving'' \\
Q3 & Primary & Multi-step reasoning & ``Large Language Models'' AND ``Mathematical Reasoning'' AND ``Chain of Thought'' \\
Q4 & Primary & Parameter-efficient reasoning models & ``Low Rank Adaptation'' AND ``Large Language Models'' \\
Q5 & Primary & Optimization-based reasoning & ``Reinforcement Learning'' AND ``Policy Optimization'' AND ``Large Language Models'' \\
Q6 & Secondary & Datasets and benchmarks (RQ1) & (``Large Language Models'' OR ``LLMs'') AND (``Mathematical Reasoning'' OR ``Math Word Problems'' OR ``Theorem Proving'') AND (``Benchmark'' OR ``Dataset'' OR ``GSM8K'' OR ``MATH dataset'' OR ``OlympiadBench'') \\
Q7 & Secondary & Modeling, prompting, and training (RQ2) & (``Large Language Models'' OR ``LLMs'') AND (``Mathematical Reasoning'' OR ``Quantitative Reasoning'') AND (``Chain-of-Thought'' OR ``Tool-Integrated'' OR ``Process Reward Model'' OR ``RLHF'' OR ``Distillation'' OR ``Mixture-of-Experts'') \\
Q8 & Secondary & Faithfulness and reasoning mechanisms (RQ3) & (``Large Language Models'' OR ``LLMs'') AND (``Mathematical Reasoning'' OR ``Multi-step Reasoning'') AND (``Faithfulness'' OR ``Hallucination'' OR ``Reasoning Trace'' OR ``Step-level'' OR ``Interpretability'') \\
Q9 & Secondary & Evaluation methodology and formal verification (RQ4) & (``Large Language Models'' OR ``LLMs'') AND (``Mathematical Reasoning'' OR ``Theorem Proving'' OR ``Formal Proof'') AND (``Evaluation Metric'' OR ``Verification'' OR ``Monte-Carlo Search'' OR ``Exact Match'' OR ``False Positives'') \\
Q10 & Secondary & Limitations and generalization (RQ5) & (``Large Language Models'' OR ``LLMs'') AND ``Mathematical Reasoning'' AND (``Generalization'' OR ``Data Contamination'' OR ``Memorization'' OR ``Out-of-distribution'' OR ``Robustness'') \\
\bottomrule
\end{tabularx}
\end{table}

\subsection{Identification, Deduplication, and Screening}
We retrieved 106,900 indexed hits across the five information sources. Because several databases do not support complete bulk export and lower-ranked results rapidly become topically irrelevant, we adopt a relevance-ranked retrieval strategy. This process retains 8,527 records for screening. After title, DOI, and manual duplicate checking, we remove 976 duplicate records, leaving 7,551 unique publications for title screening.

\begin{figure}[t]
\centering
\includegraphics[width=0.88\linewidth]{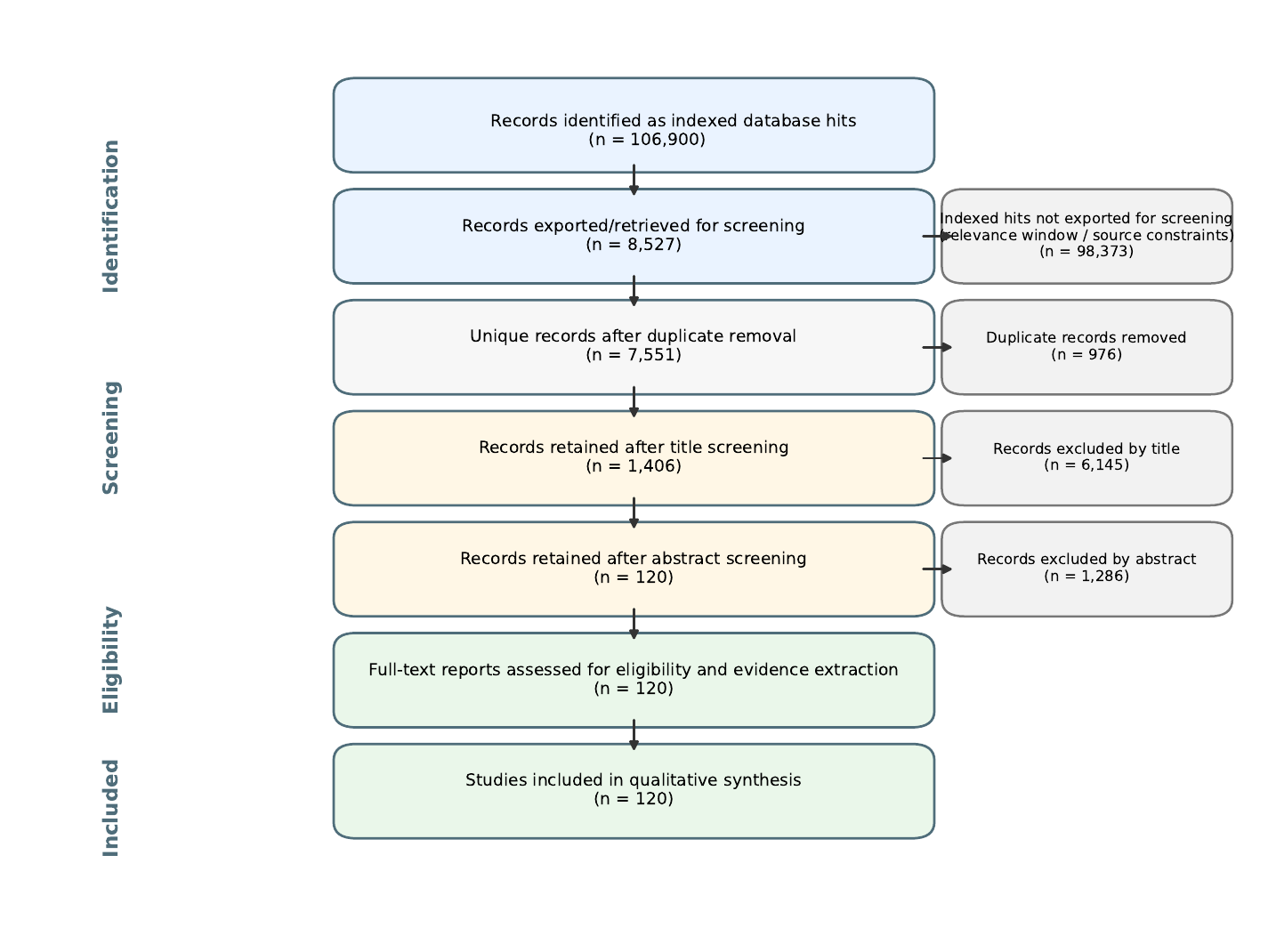}
\caption{Identification and screening flow.}
\label{fig:prisma-flow}
\end{figure}

We retain 1,406 records after title screening and exclude 6,145 records that are generic, off-topic, duplicated, or outside the scope of mathematical reasoning. Abstract screening further retains 120 studies and excludes 1,286 records that do not satisfy the inclusion criteria or lack sufficient methodological relevance. Because the search log does not record separate full-text exclusions after abstract screening, Figure~\ref{fig:prisma-flow} conservatively reports 120 full-text reports assessed and 120 studies included in the qualitative synthesis.

\begin{table}[t]
\centering
\small
\caption{Search, screening, and inclusion counts.}
\label{tab:screening-counts}
\begin{tabularx}{\linewidth}{p{0.28\linewidth}p{0.12\linewidth}X}
\toprule
Stage & Count & Notes \\
\midrule
Database identification & 106,900 & Indexed hits returned across the five logged sources. \\
Structured retrieval/export & 8,527 & Records exported or otherwise retained for screening using relevance-ranked retrieval windows. \\
Not exported for screening & 98,373 & Indexed hits outside the extraction window or not practically exportable; reported for transparency, not screened individually. \\
Duplicates removed & 976 & Computed as retrieved records minus unique records: 8,527 - 7,551. \\
Unique records screened by title & 7,551 & Deduplicated screening set. \\
Records excluded by title & 6,145 & Off-topic, generic LLM/NLP, non-mathematical, or otherwise outside scope. \\
Records retained after title screening & 1,406 & Advanced to abstract screening. \\
Records excluded by abstract & 1,286 & Failed inclusion criteria or lacked sufficient mathematical reasoning content. \\
Records retained after abstract screening & 120 & Advanced to full-text eligibility assessment and evidence extraction. \\
Studies included in qualitative synthesis & 120 & Final evidence base reported in the survey. \\
\bottomrule
\end{tabularx}
\end{table}

\subsection{Data Extraction}
We extract structured evidence from each included study using a predefined extraction schema. The extracted information covers bibliographic metadata, methodological characteristics, benchmark usage, model and training details, evaluation methodologies, and reported limitations. We allow a study to contribute to multiple research questions whenever it addresses more than one aspect of mathematical reasoning.

\begin{table}[t]
\centering
\small
\caption{Evidence extraction schema used for thematic synthesis.}
\label{tab:extraction-schema}
\begin{tabularx}{\linewidth}{p{0.25\linewidth}X}
\toprule
Extraction dimension & Fields extracted \\
\midrule
Bibliographic metadata & Authors, year, venue, publication type, DOI/arXiv URL where available, and source database. \\
RQ mapping & Primary research question(s) addressed: datasets/benchmarks, architectures/training, faithfulness, evaluation, or limitations/generalization. \\
Task and benchmark information & Reasoning task type, dataset/benchmark name, difficulty level, answer format, and evaluation setting. \\
Model and method information & Base model, architecture class, training strategy, prompting/inference mechanism, tool use, verifier, reward model, or PEFT method. \\
Evaluation information & Metrics, answer-level versus process-level assessment, symbolic/formal verification, robustness testing, and contamination analysis. \\
Critical observations & Reported limitations, failure modes, reproducibility notes, generalization findings, and open challenges. \\
\bottomrule
\end{tabularx}
\end{table}

\subsection{Synthesis Strategy}
We adopt qualitative thematic synthesis instead of quantitative meta-analysis because mathematical reasoning studies differ substantially in model scale, prompting strategy, decoding method, benchmark version, tool integration, verifier design, training procedure, and evaluation protocol. Therefore, a direct comparison of reported accuracy provides limited insight into underlying reasoning capabilities.

We organize the extracted studies into five synthesis tracks corresponding to the structure of this survey: datasets and benchmarks; model development, training, and reasoning-enhancement strategies; reasoning faithfulness; evaluation methodologies; and limitations, open challenges, and future directions. Within each track, we compare studies according to three complementary dimensions: their role in the reasoning pipeline, the type of mathematical reasoning they address, and the reliability of their evaluation methodology.

This synthesis strategy supports the central argument of this survey: improvements in mathematical reasoning may arise from different mechanisms, including dataset exposure, model adaptation, reasoning elicitation, search, tool execution, verification, or benchmark familiarity. Consequently, we interpret reported improvements as systems-level effects rather than as direct evidence of intrinsic reasoning capability.

\subsection{Bias Mitigation and Reproducibility}
We adopt several measures to improve reproducibility and reduce selection bias throughout the review process. We conduct all searches within a fixed two-day time window and record the search queries, information sources, indexed hits, retrieved records, and screening decisions. We design overlapping primary and research-question-specific query families to reduce the likelihood of missing studies that use different terminology. We remove duplicate records through title matching, DOI comparison (when available), and manual verification. Finally, we define the inclusion and exclusion criteria before screening and apply them consistently during title, abstract, and full-text assessment.

We also acknowledge several limitations of the review methodology. Search platforms differ in indexing, ranking, and export functionality; for example, Google Scholar does not fully support reproducible bulk export. Consequently, we adopt a relevance-window retrieval strategy that provides transparent, although not exhaustive, coverage of the literature. In addition, mathematical reasoning in LLMs evolves rapidly through new pre-prints, benchmark releases, and model updates. Studies published after the search period are therefore not included in this review. We report these limitations explicitly to support a transparent interpretation of the final corpus of 120 included studies.


\section{Background}\label{sec3}
We introduce the conceptual foundations of mathematical reasoning and its interaction with LLMs. We first discuss the characteristics that distinguish mathematical reasoning from general language understanding, then examine how LLMs approximate reasoning and why evaluating mathematical reasoning remains challenging.

\subsection{Mathematical Reasoning: Definitions and Scope}
Mathematical reasoning formalizes inference over symbolic structures, numerical quantities, and abstract relations. Unlike arithmetic computation, which follows fixed procedural rules, mathematical reasoning requires compositional abstraction, variable binding, constraint propagation, multi-step deduction, and often proof construction \citep{yuan2023well}. These characteristics make mathematical reasoning a stringent test case for generalization and logical consistency.

In LLM evaluation, mathematical reasoning spans multiple forms: arithmetic manipulation, algebraic transformation, word-problem solving, and formal theorem proving. All these forms differ in representation and complexity and require logically consistent intermediate reasoning.

\subsection{Reasoning Mechanisms in LLMs}
Most LLMs are auto-regressive and trained on large-scale textual corpora for next-token prediction \citep{wei2022chain}. Extensive pre-training of LLMs helps them acquire strong linguistic mappings and somehow numerical reasoning \citep{matzakos2023learning}; however, the maximum-likelihood training objective optimizes local token prediction only. Mathematical reasoning, on the other hand, requires globally valid logical derivations, rigorous theorem proving, and complex problem-solving \citep{xin2024advancing, forootani2025survey}. In hybrid reasoning tasks, models must jointly perform semantic parsing and symbolic transformation by identifying operators, binding operands, mapping language to formal structures, and executing multi-step derivations \citep{lample2019deep, kukreja2022machine, liu2023mathematical, fu2025improving}. Purely parametric models often entangle these processes, leading to operator misuse, arithmetic inconsistencies, and cascading reasoning errors in complex problems \citep{li2025Optimizing}. We observe that next-token prediction encourages pattern imitation rather than principled derivation, lacking true cognitive adaptability and iterative refinement \citep{duan2020machine, didolkar2024metacognitive, tian2024toward}. We do this to explain how it contributes to memorization effects and discontinuous performance across problem types and difficulty levels \citep{ahn2024large, huang2025math}.

\subsection{Prompting and Tool-Augmented Reasoning}
Recent reasoning-enhancement techniques partially mitigate these limitations. CoT prompting draws out intermediate reasoning steps and improves performance on multi-step problems \citep{wei2022chain}. Self-consistency decoding \citep{wang2023selfconsistency} and variants of structured prompting further enhance robustness \citep{aali2025structured}. TIR augments language models with external symbolic engines such as calculators or theorem provers, enabling reliable algebraic interpretation and theorem-level reasoning \citep{li2024enhancing, imani2023mathprompter}.

Despite these advances, mathematical problem solving remains a challenge in LLMs' reasoning \citep{feigenbaum1963computers, hosseini2014learning}. Even state-of-the-art models struggle on competition- and university-level benchmarks that require abstraction, deep knowledge of formulas, and non-standard solution strategies \citep{lewkowycz2022solving, chernyshev2024u, Liu2023GuidingMath}. We note that many reported improvements reflect better elicitation of reasoning rather than fundamentally stronger internal reasoning capabilities.

Existing research on mathematical reasoning in LLMs is shaped not by a single reasoning mechanism but by a family of prompting, search, tool use, and verification strategies that differ in both capability and reliability. Table~\ref{tab:reasoning_methods} summarizes the major reasoning-enhancement approaches and compares them in terms of their core idea, practical strengths, and key limitations.

\begin{table}[t]
\centering
\caption{Comparison of major reasoning-enhancement strategies for mathematical problem solving in LLMs.}
\label{tab:reasoning_methods}
\scriptsize
\resizebox{\textwidth}{!}{
\begin{tabularx}{\textwidth}{
>{\raggedright\arraybackslash}X
>{\raggedright\arraybackslash}X
>{\raggedright\arraybackslash}X
>{\raggedright\arraybackslash}X}
\toprule
\textbf{Method} & \textbf{Core Idea} & \textbf{Strength} & \textbf{Main Limitation} \\
\midrule
Chain-of-Thought (CoT) Prompting & Elicits intermediate reasoning steps before the final answer & Improves multi-step reasoning and interpretability of outputs & Generated reasoning may be plausible but unfaithful \\ \midrule
Self-Consistency Decoding & Samples multiple reasoning chains and selects the most consistent answer & Increases robustness and reduces single-path errors & Computationally expensive at inference time \\ \midrule
Tool-Integrated Reasoning (TIR) & Delegates calculation or symbolic operations to external tools & Improves computational precision and symbolic reliability & Requires orchestration, tool access, and error handling \\ \midrule
Program-of-Thought / Code-Augmented Reasoning & Converts parts of reasoning into executable code or symbolic programs & Better for arithmetic execution and compositional decomposition & Still depends on correct decomposition and tool execution \\ \midrule
Verifier-Guided Reasoning & Uses a separate verifier or reward signal to score solution paths & Encourages correctness beyond fluency & Verifier quality becomes a new bottleneck \\ \midrule
Process Supervision & Supervises intermediate reasoning steps rather than final answers only & Better alignment with step-level reasoning quality & High annotation and verification cost \\ \midrule
Retrieval-Augmented Reasoning & Retrieves formulas, proofs, or worked examples during problem solving & Useful when external mathematical knowledge is required & May improve recall more than genuine reasoning \\
\bottomrule
\end{tabularx}
}
\end{table}

As the comparison indicates, improvements in mathematical performance often arise from better elicitation, sampling, decomposition, or verification rather than from fundamentally solved reasoning. This distinction is crucial because a method may improve final-answer accuracy while still leaving unresolved questions about faithfulness, internal computation, and robustness. These issues motivate the evaluation challenges discussed next. Although these methods improve mathematical performance, evaluating whether they improve reasoning remains a separate challenge.

\subsection{Foundations of Evaluation and Faithfulness}
Evaluating mathematical reasoning introduces additional challenges. Many benchmarks emphasize final answer accuracy without verifying the correctness of intermediate reasoning steps \citep{mishra2022lila}. This evaluation paradigm makes it difficult to distinguish structured reasoning from pattern matching over dataset regularities.

Meta-cognitive and role-based prompting techniques attempt to improve self-reflection and error correction \citep{didolkar2024metacognitive, han2024rethinking, Wang2023MetacognitivePI}. However, generated reasoning traces may not faithfully represent the model's internal inference process \citep{didolkar2024metacognitive}. The gap between plausible explanations and verifiable derivations, therefore, remains significant. Furthermore, feedback-driven refinement in LLMs has not yet matched the structured guidance provided by human instruction \citep{anton2025human}.

Overall, LLM-based mathematical reasoning emerges from statistical sequence modeling rather than explicit symbolic computation. While prompting, search strategies, and tool integration can partially mitigate these limitations, they do not fully resolve the tension between probabilistic language modeling and the symbolic precision required for mathematics. 

Existing datasets and benchmarks target different forms of mathematical reasoning rather than a single reasoning capability. Therefore, distinguishing these reasoning categories provides the conceptual basis for interpreting the benchmark datasets discussed in the next section. Although existing benchmarks differ substantially in scope and difficulty, they can be organized according to the underlying reasoning operations required to solve the task.

Mathematical reasoning tasks addressed by LLMs can be broadly categorized according to the cognitive operations required to reach a solution. Prior work commonly identifies five major categories: arithmetic reasoning, equation and algebraic reasoning, word problem reasoning, symbolic or formal reasoning, and algorithmic reasoning. Table~\ref{tab:reasoning_taxonomy} organizes mathematical reasoning tasks according to their underlying reasoning operations and summarizes their representative benchmarks. Although these categories often overlap in practice, they capture the diversity of mathematical reasoning tasks currently studied in LLM research.

\begin{table}[t]
\centering
\caption{Taxonomy of mathematical reasoning tasks studied in LLM research.}
\label{tab:reasoning_taxonomy}
\scriptsize
\begin{tabularx}{0.85\textwidth}{>{\raggedright\arraybackslash}X >{\raggedright\arraybackslash}X >{\raggedright\arraybackslash}X}
\toprule
\textbf{Reasoning Type} & \textbf{Description} & \textbf{Typical Benchmarks} \\
\midrule
Arithmetic reasoning & Multi-step numerical computation & GSM8K, MAWPS \\
Equation reasoning & Algebraic manipulation and equation solving & ASDiv, MATH \\
Word problem reasoning & Mapping natural language to equations & GSM8K, MathQA \\
Symbolic/ proof reasoning & Formal derivations and theorem proving & MiniF2F, ProofNet \\
Algorithmic reasoning & Procedural or program-like reasoning steps & MATH, OlympiadBench \\
\bottomrule
\end{tabularx}
\end{table}

\section{Data Benchmarks}\label{sec4}
Progress in mathematical reasoning depends critically on the availability and quality of mathematical datasets. Also, dataset design determines what models learn, how they generalize, and which forms of reasoning they can reliably perform. The existing mathematical datasets span diverse problem types, cover multiple levels of cognitive complexity, and ideally provide intermediate reasoning traces that expose latent solution structure \citep{gong2023survey, karra2023impact}.

\begin{table}[t]
\centering
\caption{Taxonomy of mathematical datasets for training LLMs.}
\label{tab:tbl5a}
\scriptsize

\resizebox{\textwidth}{!}{
\begin{tabularx}{\textwidth}{
>{\raggedright\arraybackslash}X
>{\raggedright\arraybackslash}X
>{\raggedright\arraybackslash}X
>{\raggedright\arraybackslash}X}

\toprule
\textbf{Category} & \textbf{Corpus} & \textbf{Domain Focus} & \textbf{Purpose} \\
\midrule
Foundational pretraining & OpenWebMath, Proof-Pile & Large-scale web mathematics corpora; informal and formal texts & Domain-adaptive pretraining for mathematical syntax and notation learning. \\ \midrule
Programmatic pretraining & AMPS & Automatically generated arithmetic and algebra problems & Learning fundamental rules and intermediate reasoning patterns. \\ \midrule
Grade-School SFT & GSM8K (Train), MAWPS (Train), Camel-AI Math & Multi-step arithmetic and algebra word problems & Training step-by-step natural language reasoning using CoT signals. \\ \midrule
Advanced/ Competition SFT & MATH (Train), OpenMathInstruct-1 & Algebra, geometry, calculus, and number theory problems & Fine-tuning models for advanced symbolic reasoning and structured problem solving. \\ \midrule
Formal Reasoning Training & CASIA-PSGPS9K & Geometry-based formal reasoning datasets & Training models for theorem proving and structured logical derivations. \\
\bottomrule
\end{tabularx}
}
\end{table}

\subsection{Datasets}
Mathematical corpora also vary along several dimensions, including cognitive depth, supervision format, symbolic structure, and evaluation protocol. Consequently, mathematical benchmarks cannot be treated as homogeneous natural language resources; they must be analyzed as structured datasets that support different stages of learning and evaluation.

We organize the discussion around four complementary aspects. We first categorize datasets according to their cognitive complexity and functional role in training or evaluation. We then examine semantic and structural representations that map natural language problems to formal mathematical objects. Next, we discuss symbolic retrieval and tokenization mechanisms that influence how models encode and manipulate mathematical expressions. Finally, we analyze the major limitations of the current benchmark design.

\begin{table}[t]
\centering
\caption{Benchmarks for evaluating mathematical reasoning in LLMs.}
\label{tab:tbl5b}
\scriptsize

\resizebox{\textwidth}{!}{
\begin{tabularx}{\textwidth}{
>{\raggedright\arraybackslash}X
>{\raggedright\arraybackslash}X
>{\raggedright\arraybackslash}X
>{\raggedright\arraybackslash}X}

\toprule
\textbf{Category} & \textbf{Corpus} & \textbf{Domain Focus} & \textbf{Purpose} \\
\midrule
Grade-School Reasoning & MAWPS, ASDiv, GSM8K (Test), MATH401 & Arithmetic and algebraic word problems & Evaluating multi-step numerical reasoning and equation mapping ability. \\ \midrule
Competition-Level Reasoning & MATH (Test), OlympiadBench, OMNI-MATH, MathOdyssey & Olympiad, and high-school competition problems & Assessing deep conceptual reasoning and symbolic manipulation. \\ \midrule
University-Level Problems & OCW Course, U-Math & Undergraduate mathematical assignments, & Evaluating higher-education level reasoning and abstract mathematical concepts. \\ \midrule
Formalization and Theorem Proving & MiniF2F, Geometry3K, ProofNet, Def\_Wiki, Def\_ArXiv & Formal theorem proving and autoformalization tasks & Testing symbolic reasoning and formal logic verification. \\ \midrule
Generalization and Robustness & LILA, Math-Perturb & Multi-task and adversarial math reasoning tasks & Evaluating robustness and out-of-distribution generalization. \\
\bottomrule
\end{tabularx}
}
\end{table}

Tables~\ref{tab:tbl5a} and \ref{tab:tbl5b} organize existing datasets according to their roles in training and evaluation. Table~\ref{tab:benchmark_compare} complements this taxonomy by comparing widely used benchmarks in terms of reasoning type, difficulty, answer format, and evaluation characteristics. 

\begin{table}[t]
\centering
\caption{Comparative overview of major benchmarks for mathematical reasoning in LLMs.}
\label{tab:benchmark_compare}
\scriptsize
\resizebox{0.95\textwidth}{!}{
\begin{tabularx}{0.95\textwidth}{
>{\raggedright\arraybackslash}X
>{\raggedright\arraybackslash}X
>{\raggedright\arraybackslash}X
>{\raggedright\arraybackslash}X
>{\raggedright\arraybackslash}X
>{\raggedright\arraybackslash}X}
\toprule
\textbf{Benchmark} & \textbf{Reasoning Type} & \textbf{Difficulty} & \textbf{Answer Format} & \textbf{Strength} & \textbf{Limitation} \\
\midrule
GSM8K & Multi-step arithmetic word problems & Grade school & Numeric answer with rationale & Standard benchmark for stepwise arithmetic reasoning & Limited conceptual diversity; template bias \\ \midrule

MAWPS/ ASDiv & Arithmetic and algebra word problems & Elementary middle school & Equation or numeric answer & Useful for equation mapping and semantic parsing & Small scale; lower difficulty \\ \midrule

MATH & Symbolic problems across algebra, geometry, calculus, number theory & Competition/ Olympiad & Final boxed answer with solution trace & Strong benchmark for symbolic reasoning & Final-answer scoring ignores reasoning validity \\ \midrule

Olympiad Bench/ OmniMATH/ MathOdyssey & Olympiad style problems & High school Olympiad+ & Free form solution or final answer & Tests deep abstraction and non-routine reasoning & Difficult to evaluate process-level reasoning \\ \midrule

U-Math/ OCW sets & University level mathematics & Undergraduate & Free-form derivation or final answer & Reflects higher-level mathematical reasoning & Small datasets; weak standardization \\ \midrule

MiniF2F/ ProofNet/ Geometry3K & Formal proof and theorem proving & Advanced symbolic reasoning & Formal proof object & Enables rigorous symbolic verification & Narrow scope compared to word problems \\ \midrule

LILA/ Math-Perturb & Generalization and robustness tests & Mixed & Task-dependent & Useful for robustness and OOD evaluation & Not a comprehensive reasoning benchmark \\

\bottomrule
\end{tabularx}
}
\end{table}

This comparison shows that the choice of benchmark strongly shapes the model's reasoning capability. Benchmarks such as GSM8K emphasize multi-step numerical reasoning \citep{cobbe2021training}, whereas MATH \citep{hendrycks2021measuring} and Olympiad-style corpora probe deeper abstraction and symbolic manipulation \citep{toshniwal2024openmath, paster2023openwebmath}. In contrast, formal proof datasets \citep{zhang2024fuse} prioritize verifiability and logical rigor. These differences reinforce the need to examine not only dataset content but also the structural representations through which mathematical problems are encoded and solved.

\subsection{Semantic and Structural Representations}\label{sec4.2}
In mathematical reasoning, textual descriptions become insufficient for representing mathematical structure as complexity increases. Mathematical problems typically involve operator hierarchies, symbolic dependencies, variable binding, and latent quantitative relations, for which plain text representations often remain ambiguous \citep{shalyt2025llms}. Therefore, many mathematical datasets incorporate semantic or structural representations that map problem statements to formal mathematical objects.

Semantic parsing converts textual descriptions into executable expressions or formal logical structures, enabling equation construction and symbolic reasoning \citep{roy2015solving}. Unit dependency graphs represent relationships among quantities, units, and operations, helping models to infer equation structure and resolve numerical references \citep{roy2017unit, dave2024investigating}. Domain-adapted encoders such as MathBERT further support this process by learning symbol-aware representations from math-rich corpora containing LaTeX expressions and structured notation \citep{peng2023mathbert}. Contextual embedding and in-context learning techniques \citep{devlin2019bert, brown2020language} complement them by allowing models to infer latent schemas from demonstrations and better interpret variables, operators, and mathematical relations.

Together, these approaches advance mathematical datasets from surface-level text representations to structured semantic supervision. However, semantic representations alone do not complete the reasoning pipeline. Mathematical datasets must also support the retrieval, encoding, and manipulation of symbolic expressions during reasoning.

\subsection{Symbolic Representation and Retrieval}\label{sec4.3}

After identifying the semantic structure, reasoning models need to retrieve, encode, and manipulate formal symbolic objects. This step is particularly important in theorem proving, proof synthesis, and auto-formalization tasks, where reasoning depends on operations over structured expressions \citep{stathopoulos2023retrieval}. Formal representation standards such as MathML address this challenge by separating visual layout from semantic structure. Presentation MathML captures the two-dimensional rendering of expressions, while Content MathML encodes operator–argument relationships and functional composition \citep{ausbrooks2003mathml, zanibbi2012recognition}. Building on these representations, Symbol Layout Trees and Operator Trees model spatial and logical expression structure, enabling indexing, normalization, and structural matching of mathematical expressions.

Hybrid retrieval frameworks like MCAT combine textual and symbolic views to support structured math retrieval \citep{yoko2014mcat, kristianto2016mcat}. Dense retrieval approaches such as Tangent-CFT extend this by learning similarity functions over symbolic paths and expression structures \citep{mansouri2019tangent}. These methods collectively shift reasoning systems from token-level processing to structure-aware retrieval and manipulation of mathematical expressions.

\subsection{Tokenization Mechanism}\label{sec4.4}
Despite advances in semantic parsing and symbolic retrieval, numerical tokenization remains a persistent bottleneck for mathematical reasoning in LLMs. This is because LLM tokenizers are designed for natural language generation and frequently fragment numbers in ways that disrupt place-value structure and weaken arithmetic reasoning.

For example, Byte Pair Encoding (BPE) remains widely used but often splits numbers inconsistently and fails to preserve numerical regularities \citep{zouhar2023formal}. Alternative schemes explicitly encode positional structure. Left-to-right tokenization processes digits from the most significant position, whereas Right-to-left tokenization aligns tokenization with carry-based arithmetic operations \citep{singh2024tokenization}. Digit-level tokenization preserves exact place-value information but increases sequence length \citep{aakanksha2023palm}. Continuous numerical encodings such as xVal bypass discrete tokenization and embed magnitude directly in continuous space \citep{golkar2023xval}.

These observations show that mathematical reasoning depends not only on model and training data but also on the representational structures introduced during tokenization.

\subsection{Data Contamination}
Recent studies show that the structural vulnerabilities and reasoning failures observed in LLMs are inseparable from the problem of data contamination. Data contamination occurs when evaluation datasets are accidentally leaked into pretraining or fine-tuning data, enabling models to achieve high benchmark accuracy through superficial pattern matching rather than actual logical inference. Such leakages fundamentally distort learning dynamics, creating an illusion of reasoning that shatters under dynamic evaluation.

\subsubsection{Optimistic Bias}
Performance variance on perturbed benchmarks offers a reliable statistical indicator of data contamination \citep{mirzadeh2025gsm}. Frameworks like GSM-SYMBOLIC reveal that an LLM's reasoning ability is not fixed but varies widely across different symbolic versions of the same problem template \citep{mirzadeh2025gsm}. Notably, a model's nominal accuracy on the static GSM8K benchmark often lies at the extreme right tail of its perturbed performance distribution \citep{mirzadeh2025gsm}. This pronounced statistical anomaly reveals an ``optimistic bias'', indicating that static test examples have leaked into training, thereby inflating evaluation metrics and masking the model's true generalization limitations in mathematical reasoning.

\subsubsection{Memory Dependence}
The extent of contamination becomes evident when models face incomplete or mathematically impossible queries. In Math-RoB-Delete evaluations, where key constraints are removed to make problems unsolvable, contaminated models fail to apply formal logic to detect missing parameters \citep{yu2025mathrob}. Instead, they generate spurious completions by auto-regressively hallucinating the absent data \citep{yu2025mathrob}. This memory dependence directly reflects data leakage: because the prompt's semantic structure matches memorized training examples, the model skips logical inference, retrieves latent sequences from memory to fill in gaps \citep{yu2025mathrob}.

\subsubsection{False Positives}
The mechanism by which contamination undermines mathematical reasoning is revealed through model learning dynamics \citep{kang2025what}. With continued fine-tuning on contaminated data, models shift from exploring diverse reasoning paths to rigidly replicating target solutions verbatim \citep{kang2025what}. This creates a critical vulnerability that results in false positives. Contaminated models often produce correct final answers despite flawed or skipped intermediate reasoning steps. Therefore, evaluating mathematical LLMs must go beyond outcome accuracy to rigorously verify the logical validity of intermediate steps. Verbatim memorization makes models hyper-fragile, as minor numerical or semantic perturbations can cause severe drops in accuracy \citep{kang2025what}.

\subsubsection{Generalization vs. Overfitting}
Fundamentally, data contamination represents an extreme form of overfitting, in which models are trained to excel on isolated, localized test cases instead of abstracting broad mathematical principles. This reflects the central learning challenge highlighted by the Self-Error-Instruct (SEI) framework \citep{yu2025selferror}. Historically, efforts to improve reasoning by training models on isolated failure cases have failed to achieve true generalization \citep{yu2025selferror}. To overcome contamination effects, SEI emphasizes training models on generalized error typologies, such as Algebraic Errors or Conceptual Setup, clustered across diverse instances \citep{yu2025selferror}. This paradigm shift shows that mere exposure to specific, memorized ground-truth data does not foster the structural abstraction necessary for robust logical inference.

\subsection{Verifier Reliability}
\label{sec:hallucination_vectors}
Process-supervised Reward Models (PRMs) aim to address the limitations of outcome-only evaluation by scoring intermediate reasoning steps and selecting reasoning trajectories using path-aggregation techniques like MajorityVote and MinVote \citep{yu2025mathrob}. However, since PRMs share the Transformer architecture with generative LLMs, they remain vulnerable to benchmark saturation and statistical blind spots. As a result, they often encounter a ``semantic plateau'', rewarding mathematically incorrect reasoning steps that superficially mimic the linguistic patterns of correct training data.

\subsubsection{Pattern Matching vs. Verification}
The core vulnerability of PRMs lies in their reliance on probabilistic pattern-matching instead of formal logical verification \citep{mirzadeh2025gsm}. Similar to generative models that convert familiar but irrelevant semantic statements into mathematical operations \citep{mirzadeh2025gsm}, PRMs assess intermediate steps by detecting familiar syntax, LaTeX formatting, or transitional vocabulary from their training data. When an intermediate step closely aligns with these memorized templates, PRMs are prone to wrongly assigning high reward scores, effectively validating flawed logic simply because it ``looks'' correct.

\subsubsection{Path Aggregation}
Pattern-matching generators and verifiers can produce seemingly correct solutions despite flawed intermediate reasoning. Gap-filling hallucinations and verbatim repetitions mimic the linguistic structure of mathematical proofs, leading PRMs to assign them high reward scores \citep{yu2025mathrob, kang2025what}. Voting frameworks then combine these scores, often selecting reasoning paths that yield correct final answers through incorrect intermediate steps. As a result, the final answer may appear accurate even though the underlying reasoning is flawed. Thus, process-level verification does not eliminate benchmark contamination but shifts the error source from generation to evaluation.

\subsection{Benchmark Limitations}\label{sec4.5}
Current benchmark datasets continue to exhibit methodological limitations that affect how mathematical reasoning is evaluated. The previous subsections show that these limitations extend beyond individual reasoning models to the design of benchmark datasets themselves. We identify five recurring limitations in the current benchmark design.

(1) Benchmark validity remains vulnerable to training--test contamination and dataset saturation. Improving benchmark validity, therefore, requires procedurally generated benchmarks, adversarial perturbation frameworks, and continuously updated hidden evaluation sets that reduce memorization and improve evaluation reliability \citep{paster2023openwebmath}. (2) Future benchmarks also require evaluation metrics that assess both final-answer correctness and intermediate reasoning steps, reducing the influence of hallucinated reasoning and shortcut learning  \citep{tang2024mathscale, huang2024o1, guan2025rstar}. (3) Existing benchmarks systematically fail to evaluate out-of-distribution (OOD) generalization. Datasets that rely on stable problem templates encourage template matching rather than transferable mathematical abstraction \citep{lindsey2025biology}. Stress tests show that models that solve familiar formats often fail catastrophically on structurally novel variants that require the same underlying principles \citep{ahn2024large}. (4) The evaluation methodology remains highly fragmented across disconnected reasoning domains that are studied in isolation: arithmetic reasoning \citep{koncel2016mawps, miao2020diverse, cobbe2021training, yuan2023well}, competition mathematics \citep{hendrycks2021measuring, he2024olympiadbench, gao2024omni, fang2025mathodyssey}, theorem proving \citep{zheng2021minif2f, lu2021inter, azerbayev2023proofnet}, higher-education mathematics \citep{alavi2023Advances, chernyshev2024u}, and adversarial robustness. This separation makes it difficult to know whether a model has learned general mathematical reasoning or only narrow, task-specific heuristics. (5) Dataset design remains weakly aligned with the representational constraints of mathematical reasoning, particularly numerical tokenization. Structural representations and tokenization boundaries affect how models handle numerical data, yet benchmarks rarely account for these \citep{singh2024tokenization}. Overcoming these issues requires evaluation frameworks that jointly assess symbolic reasoning, semantic interpretation, OOD generalization, and tokenization effects.

\section{Architectures, Training and reasoning-enhancement Strategies}\label{sec5}
This section presents architectures, training paradigms, and reasoning-enhancement strategies for mathematical reasoning. 

\subsection{Architectures}
Recently, mathematical reasoning has moved beyond standalone language models. Current systems are combining pretrained language models with external tools, verification procedures, and planning modules, reflecting a shift from parametric reasoning to hybrid architectures that involve both generation, computation, and validation.

\textit{Language Models.}

Transformer architectures \citep{vaswani2017attention} are the foundation of modern LLMs. LLMs may be categorized into encoder-only, decoder-only, and encoder–decoder architectures based on the organization of transformer blocks. Encoder-only models such as BERT \citep{devlin2019bert} are primarily designed to understand language using bidirectional contextual representations. Decoder-only architectures, such as GPT \citep{brown2020language}, LLaMA \citep{touvron2023llama}, Qwen \citep{yang2024qwen, yang2024qwen25math}, PaLM\citep{aakanksha2023palm}, and DeepSeek \citep{liu2025deepseek}, dominate current LLM development due to their powerful auto-regressive ability. Encoder–decoder models such as T5 \citep{raffel2023exploring} integrate both text understanding and generation. Although these architectures have exhibited good performance across language tasks, their abilities are limited in mathematical reasoning because symbolic manipulation, formal verification, and multi-step computation are not encoded in their architecture. 

\textit{Tool-Augmented Architectures.}

Computational tools, such as calculators, computer algebra systems, theorem provers, symbolic solvers, and executable programming environments, are integrated into pretrained language models within tool-augmented architectures \citep{li2024enhancing}. Instead of relying solely on parametric knowledge, these systems decompose mathematical problems into subtasks and outsource specialized computations to external resources. This integration significantly enhances numerical precision and symbolic execution; however, its performance depends on the decomposition of the task and consistent collaboration between neural and symbolic components.

\textit{Verifier Architectures.}
Verifier architectures rely on dedicated modules to evaluate suggested reasoning pathways, intermediate stages, and sometimes the final answers generated by a language model \citep{cobbe2021training, wang2023selfconsistency, guo2025reward, zhang2023cumulative}. These checks often use process reward models, critic networks, or rule-based validation systems \citep{zhang-etal-2025-lessons}. 
The key factor is the separation between the component that generates an answer and the one that evaluates its correctness, which makes the reasoning seem more credible and eliminates shallow matching patterns. 
However, this approach has its limitations because the verifier's quality, accuracy, calibration, and even its approach to accepting or rejecting reasoning paths limit overall performance.

\textit{Agentic Architectures.}
Agentic architectures consider mathematical reasoning as an iterative process of decision-making, rather than a one-time generation process \citep{zhang2026systems}. These systems utilize a combination of planning, tool invocation, verification, memory, and self-correction techniques to explore multiple reasoning paths before arriving at a conclusion. Agentic systems rely on planner-executor frameworks, search-based exploration, and reflection loops to solve complex issues that, at times, require long-horizon thinking.\citep{zhang2024scalable, li2023camel}. Although these designs provide a comprehensive reasoning framework, they incur computational and coordination overheads, as well as uncertainties arising from interactions among components.
\subsection{Model Training}
Training methods significantly impact mathematical reasoning during model development. Recent approaches have adopted combining large-scale pretraining, task-specific fine-tuning, reinforcement learning, and efficient adaptation techniques to improve reasoning performance while balancing computational cost and generalization.
They often also consider distributed training strategies such as data parallelism, tensor parallelism, and pipeline parallelism, enabling efficient use of GPU/TPU clusters for training \citep{smith2022using}.
 
\textit{Pretraining.}

It provides broad linguistic and semantic knowledge by exposing models to large-scale corpora. For mathematical reasoning, domain-specific corpora such as OpenWebMath and Proof-Pile increase exposure to mathematical notation, formulas, and proofs. However, gains from mathematical pretraining must be interpreted carefully because large-scale web corpora may introduce contamination and memorization risks \citep{paster2023openwebmath, azerbayev2023proofnet, huang2025math, shi2023detecting}.

\textit{Supervised Fine-Tuning.}

It adapts pretrained models using customized datasets, such as mathematical question-answer pairs, worked examples, and reasoning traces, for task specific fine-tuning \citep{li2023label, wu2025llm, toshniwal2024openmath}.

\textit{Instruction Tuning.}

It improves language models' ability to follow natural language instructions by using curated instruction-response pairs. Mathematical instruction datasets frequently incorporate worked solutions and reasoning traces, enabling stronger generalization to multi-step problem solving \citep{min2021metaicl, parnami2022learning, chen2022large, tang2024mathscale}.

\textit{Reinforcement Learning.}

It controls reasoning behavior through reward signals. Outcome Reward Models (ORMs) \citep{thatikonda-etal-2025-logical} evaluate final answers, whereas Process Reward Models (PRMs) \citep{zhang-etal-2025-lessons} assess intermediate reasoning steps. Recent approaches further extend this paradigm through reward reasoning models and reasoning-oriented policy optimization algorithms such as Proximal Policy Optimization (PPO), Trust Region Policy Optimization (TRPO) \citep{schulman2015trust, achiam2017constrained, zhang2024scalable} and GRPO \citep{yang2024qwen25math, liu2025deepseek}.

\textit{Parameter-Efficient Fine-Tuning (PEFT).}

These methods such as LoRA, QLoRA, AdaLoRA, and DoRA reduce the computational cost of adapting large language models to mathematical tasks by updating only a small subset of model parameters while keeping most pretrained weights frozen \citep{hu2021lora, Dettmers2023QLORA, zhang2023adalora, liu2024dora}. While these methods facilitate experimentation and domain specialization, their contribution to mathematical reasoning remains largely indirect. Most reported gains arise from improved adaptation to mathematical supervision data rather than modifications to the underlying reasoning process \citep{Dou2024loramoe}. Consequently, PEFT approaches should be considered primarily as efficiency-based adaptation methods rather than as mechanisms for improving reasoning, thereby emphasizing the importance of distinguishing between adaptation efficiency and actual reasoning gains \citep{liang2025low}. 

\textit{Post-Training and Alignment.} 

Pretraining alone does not provide consistent mathematical thinking. Subsequently, post-training phases encourage a pretrained model toward organized problem solutions, improved instruction retention, and more logical reasoning alignment, or something similar. Typically, these subsequent phases combine supervised fine tuning on math data, instruction tuning in which examples demonstrate workable solutions, and reinforcement learning, which is motivated by rewards connected to results or, alternatively, to the actual process. More and more reasoning-focused models rely on post-training to improve reasoning integrity, assist with error correction, and perform solution verification in ways that pre-training alone cannot match \citep{ouyang2022training, min2021metaicl, ziegler2019fine, shao2024deepseekmath}.

\subsection{Reasoning-enhancement}
Beyond training approaches, current research has introduced a variety of methods aimed at improving reasoning at inference time. Instead of modifying model weights, like pretraining and fine-tuning do, these strategies help to direct, structure, verify, and even improve the reasoning process. Reasoning decomposition will be demonstrated using chain-of-thought prompting and search-based methodologies such as self-consistency. Then there's external tool integration and verifier-guided creation. Taken collectively, they aim to increase reasoning accuracy, make it more predictable and consistent, and enhance fidelity on difficult mathematical assignments \citep{wei2022chain, wang2023selfconsistency, li2024enhancing, cobbe2021training}.

\textit{Chain-of-Thought (CoT) Reasoning.} 

CoT prompts externalize intermediate reasoning steps before generating a final answer. This approach consistently improves performance on arithmetic and multi-step reasoning tasks by encouraging decomposition of complex problems into simpler sub-problems \citep{wei2022chain, wang2023selfconsistency, yuan2023well, parnami2022learning, chen2022large}.

\textit{Tool-Integrated Reasoning.} 

It augments language models with additional systems such as calculators, theorem provers, symbolic engines, and possibly code execution environments. In other words, they make the model less stuck and more accurate since the hard deterministic work is assigned to specialized tools, thereby increasing numerical precision and improving symbolic manipulation via delegation to those tools \citep{li2024enhancing, imani2023mathprompter, lu2021inter}.

\textit{Verifier-Guided Reasoning.} 

These techniques use auxiliary models or verification modules to evaluate possible reasoning paths and filter out faulty results. These strategies improve response dependability, but they increase reliance on verifier quality and calibration \citep{cobbe2021training, zhang2025formalizing, guo2025reward}.

\textit{Process Supervision.} 

It assesses the intermediate levels of reasoning as well as the final responses. When they provide feedback at the reasoning-process level, these techniques attempt to improve reasoning integrity and make it easier to find errors, but scale verification remains difficult \citep{zhang2025formalizing, didolkar2024metacognitive, ren2025multi}.

\subsection{Critical Synthesis: What Actually Improves Mathematical Reasoning}
Even though there has been significant development in mathematical reasoning benchmarks, the advances are mostly due to distinct processes. In other words, some strategies facilitate or direct reasoning extraction, while others improve symbolic execution, and yet others are primarily concerned with verifying outcomes and alignment. It is important to understand where each line comes from, since the increases reported in benchmarks are sometimes interpreted as proof of superior reasoning, while they may actually reflect entirely different types of capacity enhancement.

\begin{table*}[t]
\centering
\caption{Comparative synthesis of major strategies for mathematical reasoning in LLMs.}
\label{tab:math_reasoning_synthesis}
\scriptsize
\begin{tabularx}{\textwidth}{
>{\raggedright\arraybackslash}p{2.3cm}
>{\raggedright\arraybackslash}X
>{\raggedright\arraybackslash}X
>{\raggedright\arraybackslash}X}
\toprule
\textbf{Strategy} & \textbf{Why It Helps} & \textbf{Failure Mode} & \textbf{Remaining Limitations} \\
\midrule
\multicolumn{4}{c}{\textbf{Training and Adaptation}} \\
\midrule
Math Pretraining & Mathematical exposure & Contamination, memorization & Reasoning generalization \\
Instruction Tuning & Reasoning elicitation & Synthetic shortcuts & Reasoning faithfulness \\
Reinforcement Learning & Reasoning alignment & Reward hacking & Logical validity \\
PEFT & Efficient specialization & Weak reasoning effect & Reasoning mechanisms \\
\midrule
\multicolumn{4}{c}{\textbf{reasoning-enhancement}} \\
\midrule
Tool Integration & Symbolic execution & Tool failures & Intrinsic reasoning \\
Verifier Guidance & Solution filtering & Verifier hallucination & Verifier dependence \\
Process Supervision & Step-level feedback & Annotation cost & Scalable verification \\
\bottomrule
\end{tabularx}
\end{table*}

Table~\ref{tab:math_reasoning_synthesis} demonstrates that existing techniques improve mathematical reasoning through unique processes. Some strategies improve benchmark results without really strengthening the underlying argument. For example, parameter-efficient adaptation aids in the specialization of mathematical tasks, whereas tool integration delegates a portion of the reasoning process to external systems. Sure, these improvements can improve accuracy, but they may leave the fundamental reasoning restrictions unsolved.

Recent advancements demonstrate a steady trend away from parametric reasoning and toward hybrid systems that include language models, tool usage, verifiers, and structured supervision. This tendency suggests that future advancements will be determined less by model size and more by how effectively we can integrate symbolic computation, verification mechanisms, and search procedures in a single framework.

No single technique addresses all aspects of mathematical thinking. Pretraining and instruction tuning enable the model to perceive more math and provide reasoning hints, whereas tool integration facilitates symbolic execution, and verifier-guided approaches make answers appear more dependable. So, in practice, recent developments appear to be a combination of complementary processes functioning together rather than individually.

\subsection{Architectural and Training Challenges}
Despite significant improvement, numerous problems still impede valid mathematical reasoning in LLMs:

\begin{itemize} [noitemsep, topsep=0pt]
    \item \textit{Objective mismatch.}Most LLMs aim for next-token prediction rather than logically acceptable derivations, allowing fluent explanations and plausible reasoning trails to develop without ensuring mathematical accuracy \citep{lindsey2025biology}.
    \item \textit{Memorization, contamination, and benchmark validity.} Large-scale pretraining broadens exposure to mathematical material while increasing the danger of benchmark contamination, pattern recall, and data leakage. Strong benchmark performance may reflect memorization rather than true thinking skills \citep{huang2025math, shi2023detecting, feldman2020does, carlini2022privacy}.
    \item \textit{Reasoning, reliability, and hallucinations.} Models frequently generate incorrect intermediate steps, hallucinate mathematical facts, or propagate early errors through long reasoning chains. Distinguishing faithful reasoning from post-hoc rationalization remains difficult even with verifier-guided approaches \citep{ji2023survey, longpre2021entity}.
    \item \textit{Verification scalability.} Process supervision and verifier-based methods improve reasoning reliability, but scalable verification of complex reasoning trajectories remains computationally expensive and prone to verifier errors \citep{zhang2025formalizing, didolkar2024metacognitive, ren2025multi, fu2025reward}.
    \item \textit{Generalization beyond benchmark distributions.} Many models perform well on benchmark datasets yet struggle with structurally novel problems, long-horizon reasoning tasks, and out-of-distribution mathematical concepts, indicating a persistent gap between benchmark success and robust mathematical reasoning.
\end{itemize}

\section{Evaluation Frameworks}\label{sec6}

Mathematical reasoning evaluation remains challenging, as LLM outcomes may look intuitive and conversational yet contain logical or computational errors \citep{ahn2024large}. Unlike traditional text production tasks, mathematical reasoning requires organized symbolic manipulation and multi-step derivations. Observing only the final response is a weak signal, since it might mask mistakes in the chain of thought and does not demonstrate how the reasoning was built up step by step. To avoid false positives and negatives, a more reliable evaluation must evaluate both the final result and the intermediate phases \citep{wang2025examining}.

\subsection{Answer-Level Evaluation}
The primary goal of answer level evaluation is to determine whether the generated response is consistent with the reference solution. Metrics such as precision and perfect match remain popular because they are straightforward and reproducible. One may refer to it as a straightforward check because the criteria are obvious and easy to rerun. When mathematical reasoning is framed as classification, error detection, or answer verification, additional measures such as F1-score, Macro F1-score, True Positive Rate (TPR), and True Negative Rate (TNR) measure alignment between generated outputs and ground truth \citep{bostrom2020byte, huang2025math}. However, these metrics primarily evaluate the correctness of the outcome and provide limited insight into the reasoning process that produced the answer.

\subsection{Process-Level Evaluation}
Evaluation can target different levels of mathematical reasoning, ranging from final-answer correctness to stepwise derivational validity. Evaluation frameworks, therefore, differ in what they measure and what evidence they provide about reasoning quality. Table~\ref{tab:eval_metrics_math} outlines the primary evaluation methodologies and emphasizes the trade-off between simplicity, diagnostic utility, and verification rigor.

\begin{table}[t]
\centering
\caption{Major evaluation approaches for mathematical reasoning.}
\label{tab:eval_metrics_math}
\scriptsize
\renewcommand{\arraystretch}{0.95}
\begin{tabularx}{\linewidth}{
>{\raggedright\arraybackslash}p{2.5cm}
>{\raggedright\arraybackslash}X
>{\raggedright\arraybackslash}X}
\toprule
\textbf{Approach} &
\textbf{What It Evaluates} &
\textbf{Main Limitation}
\\
\midrule
Answer Accuracy
&
Final-answer correctness
&
Ignores reasoning validity
\\

Exact Match
&
Strict answer agreement
&
Sensitive to formatting and equivalent solutions
\\

Skill-Based Metrics
&
Execution of specific reasoning skills
&
Requires task decomposition
\\

Process Evaluation
&
Correctness of intermediate reasoning steps
&
Difficult to scale reliably
\\

Symbolic Verification
&
Formal validity through theorem provers or solvers
&
Limited to structured outputs
\\

Hybrid Verification
&
Neural reasoning combined with symbolic checking
&
Complex and computationally expensive
\\
\bottomrule
\end{tabularx}
\end{table}
The comparison reveals a clear tradeoff between scalability and reasoning purity. Answer-level metrics remain popular because they are easy to obtain and replicate, but they provide little insight into how the reasoning process works, such as why the model chose that response. Process-level and verification-based techniques provide more detailed diagnostic information but require organized outputs, additional supervision, or external verification mechanisms.

Recent research addresses this limitation by analyzing thinking processes. Metrics such as skill success rate, secondary skill success rate, completion rate, and calculation success rate assess the model's ability to apply specific reasoning skills to solve mathematical problems \citep{yuan2023well, didolkar2024metacognitive}. Isabelle, an automated theorem proving tool, verifies syntactic and semantic accuracy in formal mathematical reasoning tasks \citep{azerbayev2023proofnet}.

Process-level evaluation is about more than simply reviewing the result; it also involves automated process supervision, search-based error localization, and hybrid verification. With automated process supervision, you focus on the intermediate reasoning processes rather than the outcome. In other words, instead of merely asking "did it work", the system observes how it got there, which is more detailed or precise. Monte Carlo Tree Search methods provide error localization, whereas hybrid verification enhances reliability by combining symbolic solvers with executable reasoning traces \citep{ren2025multi}. However, the diversity and complexity of LLM-generated reasoning patterns make automated verification difficult \citep{herbert2019challenges, herbert2021overcoming}. Verifying intermediate reasoning steps, therefore, remains computationally expensive and often requires human oversight.

\subsection{Critical Synthesis: The Reliability Gap}
Although process-level evaluation is sometimes cited as an alternative to the shortcomings of answer-level metrics, the connection is more complicated. Answer-based metrics cannot distinguish true reasoning from memorization or shortcut learning, and process-level techniques involve a second layer of uncertainty since the accuracy of reasoning traces must be verified. As a result, assessment issues are not abolished, but rather moved from response validation to process validation.

As a result, evaluation systems must balance scalability and dependability. Answer-level metrics are simple to compute and repeat, but they provide little information about the quality of reasoning. Process-level assessment, symbolic verification, and hybrid verification provide more diagnostic information but require structured outputs, specialized tools, or an additional verifier model. This trade-off helps to explain why, despite their conceptual limitations, answer-level metrics continue to prevail when benchmarks are presented.

Mathematical reasoning assessment thus operates as a spectrum rather than a single measurement problem. distinct evaluation methodologies capture distinct elements of reasoning performance, and there is no existing paradigm that assesses correctness, fidelity, robustness, and generalization all at once.
\begin{table}[t]
\centering
\caption{Common evaluation failure modes in mathematical reasoning benchmarks.}
\label{tab:eval_failures}
\scriptsize
\begin{tabularx}{\textwidth}{
>{\raggedright\arraybackslash}X
>{\raggedright\arraybackslash}X
>{\raggedright\arraybackslash}X}
\toprule

\textbf{Failure Mode}&
\textbf{Description}&
\textbf{Implication}
\\

\midrule

Correct Answer, Incorrect Reasoning
&
The model arrives at the correct solution through flawed reasoning steps.
&
Inflates perceived reasoning ability.
\\

Wrong Answer, Mostly Correct Reasoning
&
The reasoning process is largely valid, but the final computation or formatting is incorrect.
&
Penalizes otherwise sound reasoning.
\\

Memorization
&
Correct answers result from prior exposure rather than reasoning.
&
Distorts benchmark validity.
\\

Reward Hacking
&
Models exploit weaknesses in reward functions.
&
Produces artificially inflated evaluation scores.
\\

Verifier Hallucination
&
Verifier incorrectly accepts or rejects reasoning trajectories.
&
Reduces the reliability of process supervision.
\\

Tool Invocation Failure
&
Incorrect use of external calculators or symbolic tools.
&
Introduces errors despite correct reasoning intent.
\\

Reasoning Drift
&
Early reasoning is correct, but deviates later in long solution chains.
&
Common in complex multi-step tasks. \\

\bottomrule
\end{tabularx}
\end{table}

\subsection{Evaluation Failure Modes and Limitations}\label{evalsec1}
Table~\ref{tab:eval_failures} summarizes recurring failure modes in mathematical reasoning evaluation. These failures show why current evaluation frameworks struggle to distinguish genuine reasoning from memorization, shortcut learning, verifier errors, and unstable reasoning trajectories.

These failure modes expose five broader challenges for reliable evaluation.
\begin{itemize} [noitemsep, topsep=0pt]
\item \textit{Reasoning faithfulness.}
Generated reasoning traces do not necessarily reflect the internal mechanisms that produce model outputs. As a result, seemingly coherent explanations may serve as post hoc rationalizations rather than faithful representations of the underlying reasoning process \citep{lindsey2025biology}.

\item \textit{Evaluation consistency.}
LLMs may generate different reasoning trajectories and error localization patterns across repeated executions of the same problem, complicating reproducibility and stable evaluation \citep{ren2025multi}.

\item \textit{Verifier reliability.}
Verifier-guided evaluation introduces additional dependence on evaluator quality. Verifier hallucination and reward misspecification may lead to incorrect acceptance or rejection of reasoning trajectories.

\item \textit{Scalable process verification.}
Although process-level evaluation provides richer diagnostic information than answer-level metrics, verifying intermediate reasoning steps remains computationally expensive and often requires structured outputs, external tools, or human oversight \citep{herbert2019challenges, herbert2021overcoming}.

\item \textit{Robustness and generalization.}
Current evaluation frameworks provide limited evidence regarding reasoning robustness, out-of-distribution generalization, and the ability to solve structurally novel mathematical problems.

\end{itemize}

Mathematical reasoning evaluation remains an open research problem because no single metric captures correctness, faithfulness, robustness, and generalization together. Future evaluation systems require a combination of answer-level assessment, process verification, symbolic checking, and evaluator reliability analysis to provide a more complete picture of reasoning capability.

\section{Discussion and Future Directions}\label{sec7}
The prior sections looked at datasets, reasoning structures, training methods, and evaluation as individual components. However, errors in mathematical reasoning problems are rarely generated by a single component. The reasoning pipeline must include data quality, symbolic representation, step supervision, inference, and evaluation limitations. Figure~\ref{fig:challenge_pipeline} summarizes these interactions and shows how they cumulatively hinder LLMs from reliably translating natural language into reliable symbolic thinking.

\definecolor{datacolor}{RGB}{230, 240, 255}    
\definecolor{repcolor}{RGB}{255, 245, 220}     
\definecolor{archcolor}{RGB}{230, 255, 230}    
\definecolor{traincolor}{RGB}{255, 230, 230}   
\definecolor{evalcolor}{RGB}{240, 230, 255}    
\definecolor{headerblue}{RGB}{70, 130, 180}    
\definecolor{arrowred}{RGB}{220, 20, 60}        

\tikzset{
    block/.style = {rectangle, draw=headerblue, thick, rounded corners, minimum height=5em, text width=5cm, align=center, fill=white, font=\small},
    datalist/.style = {rectangle, draw=gray, fill=datacolor, rounded corners=0.1cm, text width=4.5cm, inner ysep=3pt, font=\scriptsize, align=left},
    replist/.style = {datalist, fill=repcolor},
    archlist/.style = {datalist, fill=archcolor},
    trainlist/.style = {datalist, fill=traincolor},
    evallist/.style = {datalist, fill=evalcolor},
    outcome/.style = {trapezium, trapezium left angle=70, trapezium right angle=70, draw=arrowred, fill=arrowred!10, thick, text width=5cm, align=center, font=\small\bfseries, text=black},
    mainarrow/.style = {-{Stealth[scale=1.5]}, line width=1pt, draw=arrowred},
    connect/.style = {-{Latex}, gray, thin},
}

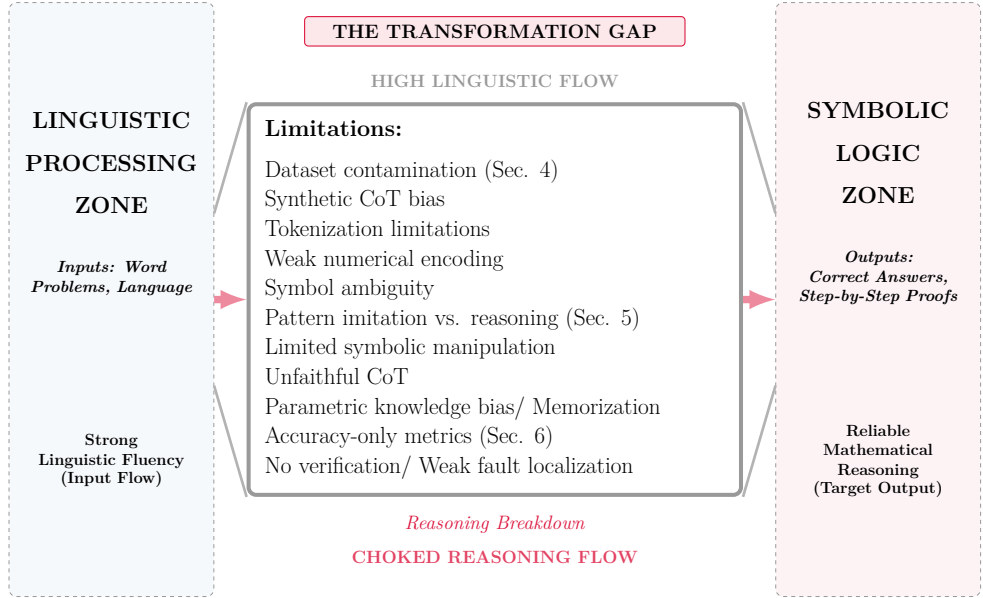
\begin{figure}[t]
\centering
\resizebox{13cm}{!}{
\begin{tikzpicture}

    \definecolor{headerblue}{RGB}{70, 130, 180}
    \definecolor{arrowred}{RGB}{220, 20, 60}

    \tikzset{
        choke/.style = {rectangle, draw=gray!80, fill=white, rounded corners=0.2cm, text width=12cm, font=\Large, align=left, inner sep=12pt, line width=3pt},
        container/.style = {rectangle, draw=black!40, dashed, rounded corners=0.2cm, minimum height=15.5cm, text width=4.5cm, inner sep=12pt, align=center, font=\small\bfseries},
        labelstyle/.style = {font=\bfseries\color{gray!80}, scale=1.1},
        gaplabel/.style = {rectangle, draw=arrowred, fill=arrowred!10, thick, text width=9.5cm, align=center, font=\large\bfseries, text=black, inner sep=6pt, rounded corners=0.1cm}
    }

    
    \node (lang_cont) [container, fill=headerblue!5] at (-2.5,0) {
        \Large LINGUISTIC \\[0.5em] PROCESSING \\[0.5em] ZONE \\[1.5em]
        \normalsize \textit{Inputs: Word Problems, Language} \\
        \vspace{3.5cm} 
        \normalsize Strong \\Linguistic Fluency \\ (Input Flow)
    };
    
    \node (bottleneck) [choke] at (7.5, 0) {
        \textbf{Limitations:} \\[0.5em]
        Dataset contamination (Sec. 4) \\
        Synthetic CoT bias \\
        Tokenization limitations \\
        Weak numerical encoding \\
        Symbol ambiguity \\
        Pattern imitation vs. reasoning (Sec. 5) \\
        Limited symbolic manipulation \\
        Unfaithful CoT \\
        Parametric knowledge bias/ Memorization \\
        Accuracy-only metrics (Sec. 6) \\
        No verification/ Weak fault localization
    };

    \node (math_cont) [container, fill=arrowred!5] at (17.5,0) {
        \Large SYMBOLIC \\[0.5em] LOGIC \\[0.5em] ZONE \\[1.5em]
        \normalsize \textit{Outputs:\\ Correct Answers,\\ Step-by-Step Proofs} \\
        \vspace{3cm}
        \normalsize Reliable\\ Mathematical Reasoning \\ (Target Output)
    };

    \begin{scope}[gray!60, line width=2pt]
        \draw (lang_cont.40) -- (bottleneck.north west);
        \draw (bottleneck.north east) -- (math_cont.140);
        
        \draw (lang_cont.-40) -- (bottleneck.south west);
        \draw (bottleneck.south east) -- (math_cont.-140);
    \end{scope}

    \node (high_flow) [labelstyle, anchor=south] at (7.5, 5.4) {HIGH LINGUISTIC FLOW};
    \node (choked_flow) [labelstyle, color=arrowred!80, anchor=north] at (7.5, -6.4) {CHOKED REASONING FLOW};

    \node (gap) [gaplabel] at (7.5, 7.0) {THE TRANSFORMATION GAP};

    \draw[->, >=latex, line width=5.5pt, draw=arrowred!50] 
        (lang_cont.east) -- (bottleneck.west);
    \draw[->, >=latex, line width=5.5pt, draw=arrowred!50] 
        (bottleneck.east) -- (math_cont.west);
        
    \node[font=\itshape\large, color=arrowred, anchor=north] at (7.5, -5.5) {Reasoning Breakdown};

\end{tikzpicture}
} 
\caption{Challenge pipeline summarizing the interconnected limitations affecting mathematical reasoning in LLMs. The “transformation gap” represents the failure to translate linguistic fluency into reliable symbolic logic due to compounding technical limitations.}
\label{fig:challenge_pipeline}
\end{figure}

\subsection{Research Trend Analysis}
The collective examination of studies reveals a clear shift in research priorities over time. Early research mainly focused on building benchmarks and prompting techniques for mathematical reasoning. As the performance of benchmarks improved, the focus gradually shifted toward enhancing the quality of reasoning through tool integration, verifier-guided inference, reinforcement learning, and process supervision. More recent work increasingly emphasizes reasoning faithfulness, benchmark contamination, process-level evaluation, agentic reasoning, and neuro-symbolic integration. It also urged the researchers to understand the model's behavior in mathematical problem solving by examining the relationship among dataset preparation, learning behavior, and the inference mechanism. Thus, there is increasing interest in investigating benchmark contamination, multilevel data leakage, memorization behavior, and their impact on model performance. Recent work further emphasizes reasoning faithfulness, process-level evaluation, agentic reasoning, and neuro-symbolic integration. In summary, this trend is a move away from maximizing benchmark accuracy and towards building mathematical reasoning systems that are more reliable, transparent, robust, and generalize beyond benchmark-specific performance.

\begin{figure}[t]
\centering
\includegraphics[width=\textwidth]{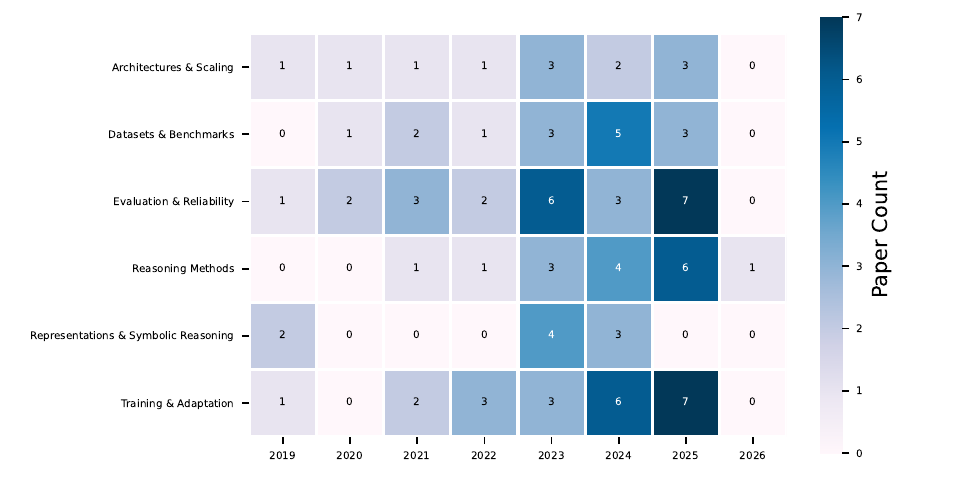}
\caption{Evolution of research focus in mathematical reasoning for LLMs. Darker cells indicate relatively greater research activity during each period. The figure summarizes broad research trends synthesized from the literature reviewed in this survey rather than quantitative publication counts.}
\label{fig:research_trends}

\end{figure}

Figure~\ref{fig:research_trends} summarizes this evolution by illustrating how research attention has shifted across the major components of the mathematical reasoning pipeline. The figure highlights that advances in mathematical reasoning no longer arise from isolated improvements in prompting or model scaling. Instead, recent research increasingly combines advances in datasets, reasoning architectures, training, and evaluation to address the limitations identified throughout this survey.

\subsection{Cross-Section Analysis}\label{dissec1}

The limitations discussed in previous sections reinforce each other rather than occurring independently. Dataset design influences the reasoning patterns learned during training, training objectives shape inference behavior, and evaluation protocols determine which behaviors are rewarded. Consequently, improvements in one component often expose weaknesses in others.

Current datasets continue to prefer benchmark results over reasoning sturdiness. Large sets like GSM8K and MATH now provide improved training resources for math reasoning. However, issues like benchmark contamination, repeated template patterns , and a lack of diversity in problems push memorization and shortcut learning rather than reasoning that can actually transfer. Hence, as benchmark accuracy grows, weaknesses become more visible, especially in symbolic handling, abstraction, and out-of-sample generalization \citep{liang2025low,cobbe2021training,hendrycks2021measuring,ji2023survey,longpre2021entity,feldman2020does}.

Training strategies have the effect of consolidating those shallow reasoning habits, and this happens regularly. While instruction tuning \citep{min2021metaicl,parnami2022learning} and reinforcement learning from human feedback (RLHF) \citep{ouyang2022training} improve instruction adherence and smoother responses, they do not necessarily improve the deeper reasoning mechanism. Models typically utilize plausible reasoning to arrive at a solution, but these stories may contradict the actual computations used \citep{lindsey2025biology}. Overall performance remains good on typical benchmark distributions, but it lowers when the math problems look structurally different, and when abstraction and long-horizon thinking are required \citep{ahn2024large,huang2023towards}.

The limitations of the evaluation protocols further obscure improvements in language model reasoning. Most benchmarks measure only the accuracy of the final answer \citep{huang2025math,rhomrasi2025llm,singh2024tokenization}, providing little information about how the solution was obtained. Consequently, logically incorrect reasoning may receive full credit if the final answer is correct, whereas mathematically sound reasoning may be penalized because of minor numerical errors \citep{yuan2023well,wang2025examining}. PRM models can also behave hallucinated during process level verification. Hence, current evaluation protocols also provide limited support for reasoning verification and fault localization \citep{ren2025multi}.

Representation remains another fundamental limitation. Mathematical reasoning requires consistent manipulation of numbers, variables, operators, and symbolic expressions. Current tokenization schemes were designed for natural language rather than mathematical structures, leading to errors in numerical encoding and symbolic reasoning \citep{rajaraman2024toward,shalyt2025llms}. Although structure-aware representations reduce some arithmetic errors \citep{singh2024tokenization}, they do not fully address failures in symbolic abstraction and mathematical manipulation. Existing evidence suggests that unit-digit tokenization consistently outperforms byte-pair tokenization in arithmetic and symbolic reasoning tasks \citep{singh2024tokenization}. However, its effectiveness tends to decline when problems simultaneously require both semantic understanding and symbolic manipulation, exposing a fundamental limitation of current representation schemes across a wide range of mathematical reasoning tasks.

These observations indicate that mathematical reasoning failures are rarely caused by a single weakness. Instead, they emerge from interactions among dataset design, model representation, training supervision, inference behavior, and evaluation methodology. Improving only one component is therefore unlikely to produce reliable mathematical reasoning.

Table~\ref{tab:failure_modes} illustrates how these interacting limitations appear during mathematical problem solving. Although they differ in their immediate causes, most originate from the same underlying weaknesses in representation, supervision, and evaluation.

\begin{table}[t]
\centering
\caption{Common failure modes in LLM-based mathematical reasoning and their broader implications.}
\label{tab:failure_modes}
\scriptsize
\resizebox{\textwidth}{!}{
\begin{tabularx}{\textwidth}{
>{\raggedright\arraybackslash}X
>{\raggedright\arraybackslash}X
>{\raggedright\arraybackslash}X
>{\raggedright\arraybackslash}X}
\toprule
\textbf{Failure Mode} & \textbf{Description} & \textbf{Likely Cause} & \textbf{Implication} \\
\midrule
Arithmetic inconsistency & Incorrect intermediate or final calculations & Weak numerical representation or carry handling & Reduces trust even in otherwise coherent solutions \\ \midrule
Operator misuse & Wrong algebraic transformation or symbolic manipulation & Poor structural grounding of expressions & Leads to logically invalid derivations \\ \midrule
Reasoning drift & Early correct reasoning followed by later off-track steps & Long-horizon dependency failure & Common in multi-step and competition-level tasks \\ \midrule
Template overfitting & Success on familiar formats but failure on novel variants & Benchmark regularity and shortcut learning & Inflates apparent reasoning ability \\ \midrule
Unfaithful chain-of-thought & Explanation does not reflect the actual internal decision process & Post hoc rationalization and fluency bias & Weakens interpretability and process trustworthiness \\ \midrule
Memorization / contamination & Correct answer due to prior exposure rather than reasoning & Overlap between training and evaluation data & Distorts benchmark validity \\ \midrule
Symbol grounding failure & Inability to reliably map text, variables, and equations & Weak semantic-to-symbolic alignment & Harms word problem solving and formalization \\ \midrule
Verification gap & Correct-looking solution cannot be reliably checked stepwise & Lack of scalable process-level evaluation & Makes evaluation incomplete and potentially misleading \\
\bottomrule
\end{tabularx}
}
\end{table}


\subsection{Research Prioritization Framework}
Table~\ref{tab:research_priorities} organizes future research directions according to their expected impact, feasibility, and anticipated research horizon. Future research directions differ in both feasibility and expected impact. Some limitations can be addressed within existing LLM training pipelines, whereas others require new reasoning architectures.

Near-term progress is likely to come from improving evaluation protocols, reducing benchmark contamination, and developing methods that encourage faithful reasoning. These problems are already well understood and can be addressed without fundamental architectural changes.

Mid-term progress will require tighter integration between language models and symbolic reasoning systems. Numerical representations, external mathematical tools, planning, verification, and self-correction must all function together rather than separately.

Long-term advancement is anticipated to rely on neuro-symbolic systems that integrate statistical language modeling with formal mathematical reasoning. These systems seek to maintain the flexibility of LLMs while offering the dependability necessary for symbolic mathematics. Overall, the data examined in this survey indicate that continuous model growth is unlikely to solve current reasoning difficulties.

\begin{table}[t]
\centering
\caption{Prioritized research directions for advancing mathematical reasoning in LLMs.}
\label{tab:research_priorities}
\scriptsize
\begin{tabularx}{\linewidth}{
>{\raggedright\arraybackslash}X
>{\centering\arraybackslash}p{1.5cm}
>{\centering\arraybackslash}p{1.5cm}
>{\centering\arraybackslash}p{1.8cm}}
\toprule
\textbf{Research Direction} &
\textbf{Impact} &
\textbf{Feasibility} &
\textbf{Horizon}
\\
\midrule
Process-level evaluation and verification
&
High
&
High
&
Near-term
\\
Benchmark contamination mitigation
&
High
&
High
&
Near-term
\\
Faithful reasoning mechanisms
&
High
&
Medium
&
Near-term
\\
Numerical and symbolic representations
&
High
&
Medium
&
Mid-term
\\
Agentic and hybrid mathematical reasoning
&
Very High
&
Medium
&
Mid-term
\\
Neuro-symbolic mathematical systems &Very High&Low&Long-term\\
\bottomrule
\end{tabularx}
\end{table}

\subsection{Open Problems and Future Directions}\label{dissec2}
The limitations identified throughout this survey point to several research directions that directly address the current weaknesses of LLM-based mathematical reasoning.

\subsubsection{Synthetic Reasoning Datasets}
Current training pipelines rely heavily on synthetic chain-of-thought data \citep{guan2025rstar}, while recent studies report diminishing returns from further dataset refinement \citep{huang2024o1,tang2024mathscale}. Increasing the amount of mathematical supervision also risks degrading general language capabilities, motivating approaches that isolate mathematical knowledge using mechanisms such as adapters \citep{Dou2024loramoe}. Future datasets should therefore emphasize human-curated reasoning, diverse solution strategies, and out-of-distribution mathematical problems that cannot be solved through memorization alone \citep{huang2025math}.

\subsubsection{Faithful Reasoning Mechanisms}
Current models often provide answers that appear logically consistent but do not represent their underlying thinking process \citep{lindsey2025biology}. Improving reasoning faithfulness requires training objectives that align provided explanations with actual model computations. Promising attempts include structured reasoning supervision, improved incentive models for reasoning \citep{mu2024rule}, and architectures designed for better interpretability.

\subsubsection{Process-Level Evaluation}
Final-answer correctness is just a partial indicator of mathematical reasoning competence \citep{yuan2023well}. Future assessment systems should validate intermediate reasoning processes, highlight areas where reasoning fails, and distinguish proper reasoning and accidental correctness. Hybrid evaluation frameworks that combine symbolic verification and neural reasoning \citep{azerbayev2023proofnet}, automated process supervision \citep{ren2025multi}, and reasoning-aware evaluation metrics provide promising directions.

\subsubsection{Agentic and Hybrid Mathematical Reasoning}
Recently, systems have steadily combined language models with planning, tool use, verification, and iterative refinement \citep{huang2025metamath,zhang2026systems}. These agentic settings address limits that remain difficult for a single-pass reasoning method, particularly in long-horizon mathematical problems. Future research should focus on dependable coordination between language models, external instruments, and symbolic checkers, while maintaining transparency and ensuring that assessments remain reliable.

\subsubsection{Numerical and Symbolic Representation}
Reliable mathematical reasoning is dependent on an accurate representation of numbers, variables, and symbolic expressions. Existing tokenization techniques frequently disregard mathematical structure, which limits symbolic manipulation \citep{rajaraman2024toward}. Future research should focus on structure-aware representations that explicitly model mathematical objects rather than treating them as ordinary text tokens. Approaches that use mathematical encoders and symbolic representations \citep{peng2023mathbert,zanibbi2012recognition} offer a potential foundation for this path.

\section{Limitations}\label{sec8.1}

Despite rigorous efforts being made to provide a comprehensive, structured synthesis of mathematical reasoning in LLMs, we acknowledge several limitations: 

\begin{itemize}[noitemsep, topsep=0pt]
\item \textit{Rapidly evolving literature.}
Research on mathematical reasoning in LLMs is rapidly evolving, with new "reasoning" models, benchmarks, and training methodologies emerging frequently. The literature collection procedure does not incorporate all subsequent publications.

\item \textit{Limited access to proprietary systems.}
The findings of this study are based on publicly available articles and technical reports. Commercial systems generally give little information about their datasets, architectures, training approaches, and assessment methods, making direct comparison difficult.

\item \textit{Heterogeneous evaluation settings.}
The study's conclusions include a range of standards, prompting strategies, model sizes, and evaluation methodologies. These distinctions immediately limit performance comparisons between models, preventing a comprehensive quantitative meta-analysis.

\item \textit{Diverse benchmark objectives.}
Existing benchmarks tend to include a wide range of mathematical reasoning methodologies, including arithmetic, algebra, geometry, theorem proof, formal verification, and even contest mathematics. This clarifies why the aims and difficulty levels are not directly equivalent.

\item \textit{Literature coverage.}
Although the survey follows a structured methodology for collecting the literature, the size and growth of the field make exhaustive coverage difficult. Some relevant studies may therefore not be included.

\end{itemize}

\section{Conclusions}\label{sec9}
Mathematical reasoning remains one of the most significant challenges for complex language models, as it requires numerical and symbolic manipulation, abstraction, multi-step deduction, and plain logical consistency, in addition to smooth natural language generation. In this study, recent advances in LLMs for arithmetic reasoning problems were examined from four perspectives: datasets and benchmarks, reasoning architectures, training and reasoning augmentation strategies, and evaluation methodologies.

The analysis demonstrates that mathematical thinking improves as a result of the interaction of several components rather than from a single architectural innovation. The quality of the dataset affects the reasoning patterns utilized during training. Although reasoning architectures contribute, particularly through symbolic computation and planning, real training procedures help guide reasoning behavior, and evaluation protocols ultimately determine which types of skills will be tested. Shortcomings in any of these components affect the correctness of mathematical reasoning.

This survey identifies many persistent obstructions that continue to restrict progress in mathematical reasoning, similar to a drag that you perceive but can't easily identify. The major factors are: benchmark contamination, synthetic reasoning bias, and the type of disloyal stream of thought that appears when understanding is lacking, both numerically and symbolically. There are various evaluation techniques that focus on whether the final response is correct. Taken together, these limits show that benchmark accuracy alone is inadequate for measuring mathematical thinking.

Future breakthroughs will need reliable reasoning mechanisms with robust process-level assessment, enhanced numerical and symbolic representations, and comprehension, with a primary emphasis on deeper integration between language models and symbolic reasoning systems. Addressing these problems would require integrated solutions in datasets, reasoning frameworks, training, and evaluation, rather than continuing model scaling alone.

{\small
\bibliographystyle{elsarticle-harv}
\bibliography{references}
}
\end{document}